\documentclass[10pt,journal]{IEEEtran}
\usepackage[table]{xcolor}
\usepackage{epsfig}
\usepackage{graphicx}
\usepackage{amsmath}
\usepackage{amssymb}
\usepackage{enumitem}
\usepackage{pifont}
\usepackage[linesnumbered,titlenumbered,ruled,vlined,commentsnumbered]{algorithm2e}
\SetKwComment{Comment}{$\triangleright$\ }{}

\usepackage{multirow}
\usepackage{tabularx}
\usepackage{tcolorbox}
\usepackage{dsfont}
\usepackage{url}
\usepackage[tight]{subfigure}
\usepackage{color}
\usepackage{subfigure}
\usepackage{amsthm}
\usepackage[export]{adjustbox}
\usepackage{comment}
\usepackage[utf8]{inputenc} 
\usepackage[T1]{fontenc}    
\usepackage{url}            
\usepackage{booktabs}       
\usepackage{amsfonts}       
\usepackage{nicefrac}       
\usepackage{microtype}      
\usepackage{diagbox}
\usepackage{balance}
\usepackage[colorlinks,linkcolor=red,citecolor=blue]{hyperref}

\usepackage{xspace}
\makeatletter
\DeclareRobustCommand\onedot{\futurelet\@let@token\@onedot}
\def\@onedot{\ifx\@let@token.\else.\null\fi\xspace}
\def\eg{\emph{e.g}\onedot} 
\def\ie{\emph{i.e}\onedot} 
 
\def\etc{\emph{etc}\onedot} 
 
\def\etal{\emph{et al}\onedot}
\makeatother

\usepackage{scalerel}
\usepackage{tikz}
\usetikzlibrary{svg.path}
\definecolor{orcidlogocol}{HTML}{A6CE39}
\tikzset{
  orcidlogo/.pic={
    \fill[orcidlogocol] svg{M256,128c0,70.7-57.3,128-128,128C57.3,256,0,198.7,0,128C0,57.3,57.3,0,128,0C198.7,0,256,57.3,256,128z};
    \fill[white] svg{M86.3,186.2H70.9V79.1h15.4v48.4V186.2z}
                 svg{M108.9,79.1h41.6c39.6,0,57,28.3,57,53.6c0,27.5-21.5,53.6-56.8,53.6h-41.8V79.1z M124.3,172.4h24.5c34.9,0,42.9-26.5,42.9-39.7c0-21.5-13.7-39.7-43.7-39.7h-23.7V172.4z}
                 svg{M88.7,56.8c0,5.5-4.5,10.1-10.1,10.1c-5.6,0-10.1-4.6-10.1-10.1c0-5.6,4.5-10.1,10.1-10.1C84.2,46.7,88.7,51.3,88.7,56.8z};
  }
}
\newcommand\orcidicon[1]{\href{https://orcid.org/#1}{\mbox{\scalerel*{
\begin{tikzpicture}[yscale=-1,transform shape]
\pic{orcidlogo};
\end{tikzpicture}
}{|}}}}

\begin{document}
%
\title{Texture Re-scalable \\ Universal Adversarial Perturbation}
%
%
%
\author{Yihao~Huang,
        Qing~Guo,
        Felix~Juefei-Xu,
        Ming~Hu,\\
        Xiaojun~Jia$^\dagger$\,,
        Xiaochun~Cao,
        Geguang~Pu
        and~Yang~Liu
\thanks{Yihao~Huang, Ming~Hu, Xiaojun~Jia and Yang~Liu are with Nanyang Technological University, Singapore. Felix~Juefei-Xu is with New York University, USA. Qing~Guo is with the Institute of High Performance Computing (IHPC) and Centre for Frontier AI Research (CFAR), Agency for Science, Technology and Research (A*STAR), Singapore. Geguang~Pu is with 1) East China Normal University and 2) Shanghai Industrial Control Safety Innovation Technology Co., LTD, China. Xiaochun Cao is with School of Cyber Science and
Technology, Shenzhen Campus, Sun Yat-sen University. \\$\dagger$ Xiaojun Jia is the corresponding author.}
}

\maketitle

\begin{abstract}
Universal adversarial perturbation (UAP), also known as image-agnostic perturbation, is a fixed perturbation map that can fool the classifier with high probabilities on \textbf{arbitrary} images, making it more practical for attacking deep models in the real world.
Previous UAP methods generate a scale-fixed and texture-fixed perturbation map for all images, which ignores the multi-scale objects in images and usually results in a low fooling ratio.
Since the widely used convolution neural networks tend to classify objects according to semantic information stored in local textures, it seems a reasonable and intuitive way to improve the UAP from the perspective of utilizing local contents effectively.
In this work, we find that the fooling ratios significantly increase when we add a constraint to encourage a small-scale UAP map and repeat it vertically and horizontally to fill the whole image domain.
To this end, we propose texture scale-constrained UAP (TSC-UAP), a simple yet effective UAP enhancement method that automatically generates UAPs with category-specific local textures that can fool deep models more easily.
Through a low-cost operation that restricts the texture scale, TSC-UAP achieves a considerable improvement in the fooling ratio and attack transferability for both data-dependent and data-free UAP methods.
Experiments conducted on four state-of-the-art UAP methods, eight popular CNN models and four classical datasets show the remarkable performance of TSC-UAP.
\end{abstract}



\section{Introduction}\label{sec:intro}
\begin{figure}[tbp]
    \centering
    \includegraphics[width=\linewidth]{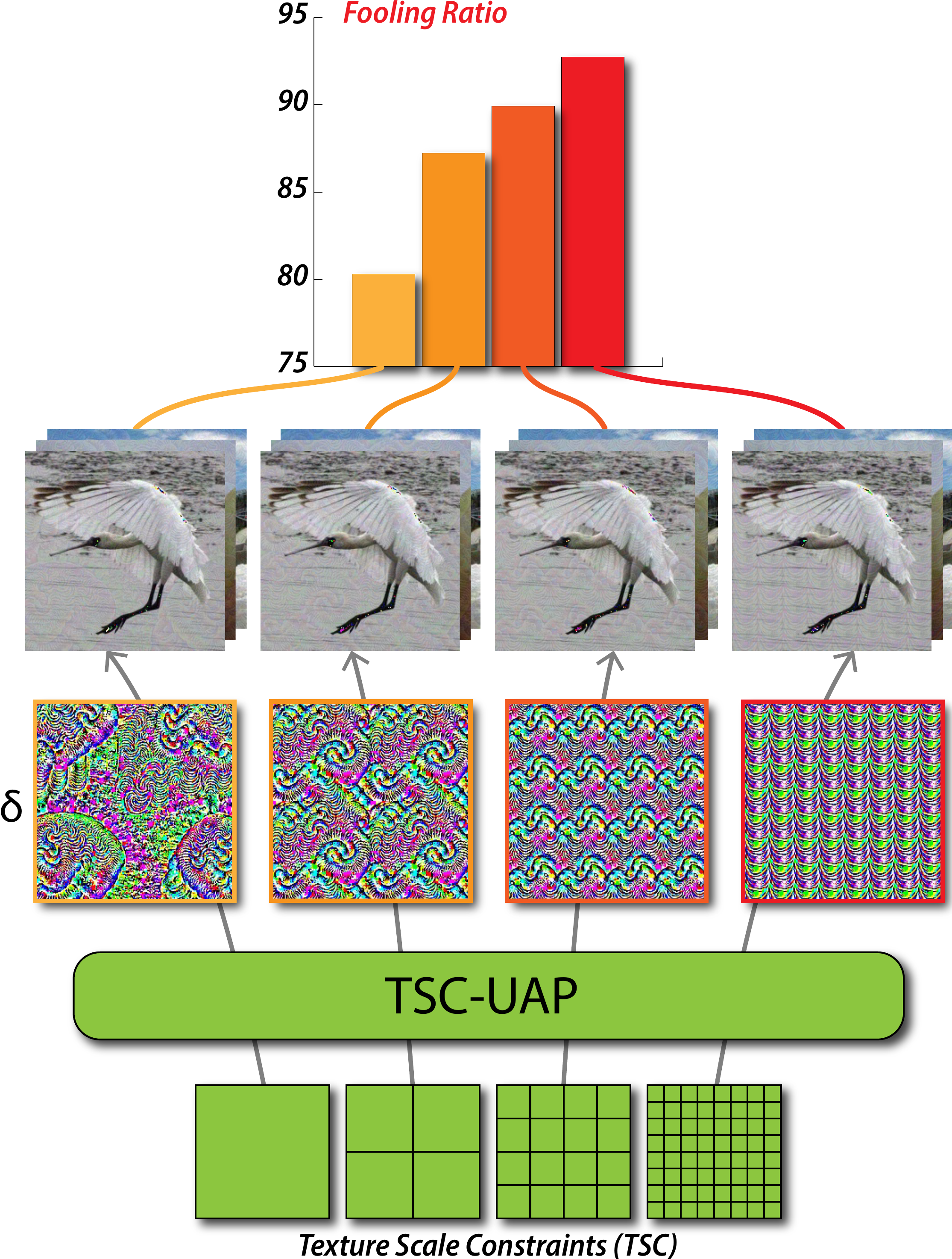}
    \caption{We propose TSC-UAP, an enhancement method that utilizes texture scale constraints to restrict the UAP. The texture scale constraints facilitate UAP to achieve higher fooling ratios, no matter data-dependent or data-free.}
    \label{fig:schematic-TSCUAP}
\end{figure}

Deep learning has shown a broad impact on computer vision tasks, including image classification \cite{he2016deep,tan2019efficientnet,zheng2022rotation,dong2022weighted,touvron2022resmlp}, object detection \cite{girshick2015fast,tan2020efficientdet,wang2023yolov7,zhou2022mmrotate,cheng2023towards}, and image segmentation \cite{he2017mask,liu2021swin,luddecke2022image,kim2022restr,ji2022amos}. However, the susceptibility of deep neural networks (DNNs) to quasi-imperceptible adversarial perturbations has raised concerns about their robustness \cite{szegedy2013intriguing,nie2022DiffPure,schlarmann2023adversarial,yang2022closer,jia2024improving,jia2024fast}. As a result, deep learning security \cite{huang2021advfilter,huang2023natural,huang2024personalization} has become a popular research area, with adversarial attacks \cite{goodfellow2015explaining,kurakin2016adversarial,carlini2017towards,croce2020reliable,bai2021transformers,FengRobust2024,GoelFast2022,WangDispersed2022,ZhaoPluggable2024,ZhangWalking2021,huang2023ala,han2023ot,gao2024boosting} playing a vital role in the community. However, these attacks are not practical for real-world online applications as they require complex optimization algorithms to generate adversarial examples for each input image, which is not general or efficient enough.

To address this issue, Dezfooli \etal{} \cite{moosavi2017universal} proposed a solution known as universal adversarial perturbations (UAPs), which are designed to misclassify a target CNN model over \textbf{any} images, referred to as image-agnostic attacks. For the sake of simplicity, we refer to this method as DeepFool-UAP, as it uses the DeepFool \cite{moosavi2016deepfool} algorithm to generate the UAPs. It is evident that the pre-computed UAPs represent a significant threat to real-time applications, which provides a major advantage over basic adversarial attacks. UAPs have also demonstrated their ability to carry out cross-model and cross-data attacks, which increases the likelihood of employing UAPs in the physical world applications. 

The following research focuses on exploring new architectures or new loss functions of the UAP generation method \cite{ijcai2021p635}. GAP \cite{poursaeed2018generative} and NAG \cite{mopuri2018nag} utilize generative adversarial networks (GAN) architecture to generate UAP, which attempts to capture the distribution of UAP for a given classifier. However, GAN is a complex architecture that is hard to train, which means the requirement of a generator itself is a drawback of these UAP generation approaches. These methods also do not obtain state-of-the-art fooling ratios. Thus our research does not focus on GAN-based UAPs. Compared to exploring new architectures, the innovation in loss functions is a simple modification towards regular UAP generation methods which can be transferred to other methods, showing better generality. DF-UAP \cite{zhang2020understanding} and Cos-UAP \cite{zhang2021data} propose special loss functions related to output logit vectors of the CNN model. They are the state-of-the-art UAP generation methods and Cos-UAP is a bit better than DF-UAP. Furthermore, due to the elaborate design of the loss functions in DF-UAP and Cos-UAP, they can successfully generate UAPs without any training samples. We call them data-free UAPs to distinguish them from data-dependent methods such as DeepFool-UAP, GAP, \etc. Although research on UAP has made good progress, we surprisingly find that rare works attempt to improve UAP from the perspective of texture. However, the most prominent features of UAPs are their special textures, which are significantly different from the adversarial noise in non-universal attacks.

We choose UAP generated from SGD-UAP \cite{shafahi2020universal} as an example to introduce such texture patterns. SGD-UAP is a representative UAP generation method with a regular pipeline and the state-of-the-art DF-UAP and Cos-UAP are utilizing the same pipeline as it (only differ in loss functions). As shown in $\delta$ of Figure~\ref{fig:schematic-TSCUAP}, the UAP is classified as `coral' by ResNet50 and the textures distinctly show the characteristic of coral. Due to this, we conduct experiments and find the ``category-specific local'' textures can benefit the attack performance of UAP. Based on the observation, we propose a simple yet effective UAP enhancement method with texture scale constraints, called TSC-UAP. It shows excellent performance at a minor cost.

To sum up, our work has the following contributions:
\begin{itemize}[itemsep=2pt,topsep=0pt,parsep=0pt]
\item We are the first to substantiate the significance of category-specific local texture as a new research direction to benefit UAP generation methods.
\item We propose TSC-UAP, a simple yet effective way to achieve a higher fooling ratio, attack transferability, and data efficiency with minor costs. It is also general enough to enhance data-dependent and data-free UAP methods.
\item The experiment conducted on four classical datasets and eight popular CNNs shows the effectiveness of TSC-UAP in generating better UAPs.
\end{itemize}

\section{Related Work}\label{sec:related_work}
\subsection{Basic Adversarial Attacks}
%
Szegedy \etal{} \cite{szegedy2013intriguing} proposed the first adversarial attack method, L-BFGS, which fools neural networks by adding quasi-imperceptible perturbations to input images. Inspired by \cite{szegedy2013intriguing}, Goodfellow \etal{} \cite{goodfellow2015explaining} proposed a simple yet famous method, the Fast Gradient Sign Method (FGSM), which aims to maximize the attack success rate with a restricted perturbation budget (\ie, epsilon ($\epsilon$)) in a single attack step. However, FGSM did not achieve a high attack success rate due to its one-step attack procedure. To improve the attack performance, Iterative FGSM (I-FGSM) \cite{kurakin2016adversarial} and Projected Gradient Descent (PGD) \cite{madry2017towards} have been proposed to perturb the image in multiple iterations. In each iteration, the images are only allowed to have added noise that is less than a fraction of epsilon.
%

Different from the FGSM-series methods, other popular attacks, such as DeepFool \cite{moosavi2016deepfool} and C\&W \cite{carlini2017towards}, approach adversarial attacks from a different perspective. They aim to minimize the perturbation magnitude under the situation that the image is misclassified. DeepFool \cite{moosavi2016deepfool} uses the decision boundaries of the target model as the gradient guidance and updates the gradient. Specifically, it chooses the direction that is orthogonal to the decision hyperplane in each attack iteration. C\&W \cite{carlini2017towards} formulates the adversarial attack as an optimization problem and proposes several objective functions to replace highly non-linear classification functions, resulting in better optimization. The generation of adversarial examples through complicated optimization in these basic adversarial attacks is less practical to apply in real-world attack scenarios.

\subsection{Universal Adversarial Attack and Defense}\label{sec:uap_attack_defense}
\noindent\textbf{Attack.} Dezfooli \etal{} \cite{moosavi2017universal} first proposed the concept of Universal Adversarial Perturbations (UAPs). UAP is
a single fixed image-agnostic adversarial noise that can fool most of the images from a data distribution with a given CNN model. They proposed iteratively crafting perturbations for training samples by using DeepFool \cite{moosavi2016deepfool} and limiting the perturbation magnitude with a sphere constraint. It is obvious that such UAP can fool images without additional online processes, which shows the practical possibility in real-world applications. 
To enhance the fooling ratio and attack efficiency, Shafahi \etal{}\cite{shafahi2020universal} proposed a Stochastic Gradient Descent UAP (SGD-UAP) attack. As the inner loop of the DeepFool attack is very time-consuming, they utilized PGD to optimize the perturbation by batches instead of a single image, which also guarantees convergence.

SV-UAP \cite{khrulkov2018art} exploited the Jacobian matrices in feature maps to calculate singular vectors for generating UAPs. The method is able to achieve more than 60\% fooling ratio with only 64 training samples, which is very data-efficient. However, since SV-UAP still requires training data, \cite{mopuri2017fast} aimed to generate UAPs without any training data, termed data-free UAPs. The data-free methods \cite{geirhos2018imagenet,liu2019universal,zhang2021data} focus on the neuron values of the target CNN model and increase the model's uncertainty by special loss function for generating UAPs. AT-UAP \cite{li2022learning} and TRM-UAP \cite{liu2023trm} are newly proposed state-of-the-art data-free methods which deliver superior attack performance. AT-UAP introduces a consistency regularizer to analyze the relationship among training data. TRM-UAP redefines the task of generating UAP as a truncated ratio optimization problem, aiming to refine the UAP generation process through the balance of positive and negative activations.

Inspired by Generative Adversarial Networks (GAN), Network for Adversary Generation (NAG) \cite{mopuri2018nag} was proposed to generate adversarial noise by simulating the distribution of UAPs. GAP \cite{poursaeed2018generative} also followed a similar idea. Since GAN-based UAP methods are complex and difficult to train, the use of GAN for attack is a drawback of UAP approaches. They also cannot achieve a high fooling ratio.

In previous works, only PD-UA \cite{liu2019universal} somewhat mentioned the influence of texture on UAP. They set a texture priori and force the UAP to have a similar style as it. However, the texture priori may be opposite to the generation direction of the UAP category. For example, they said the circle textural pattern achieved the best performance in their experiment. Circle texture priori may be suitable for generating animal-like UAPs for the eyes of animals are of the circle shape. It is not suitable for generating UAP categories that only have line contour, \eg, table, binder. Compared with PD-UA, our method focuses on restricting the scale of texture in generated UAP, which constructs non-prior class-related texture by the attack method automatically.

\noindent\textbf{Defense.}
Akhtar \etal{} \cite{akhtar2018defense} propose the first defense method against UAP. Their method uses a Perturbation Rectifying Network (PRN) as a kind of pre-processing. The PRN is trained with both real and fake perturbations that don't depend on the image. Mummadi \etal{} \cite{mummadi2019defending} demonstrate that adversarial training can effectively prevent UAPs. Shafahi \etal{} \cite{shafahi2020universal} suggest training robust models at a low cost through the optimization of a min-max problem, employing either alternating or simultaneous stochastic gradient approaches.

\begin{figure}[tbp]
    \centering
    \includegraphics[width=0.9\linewidth]{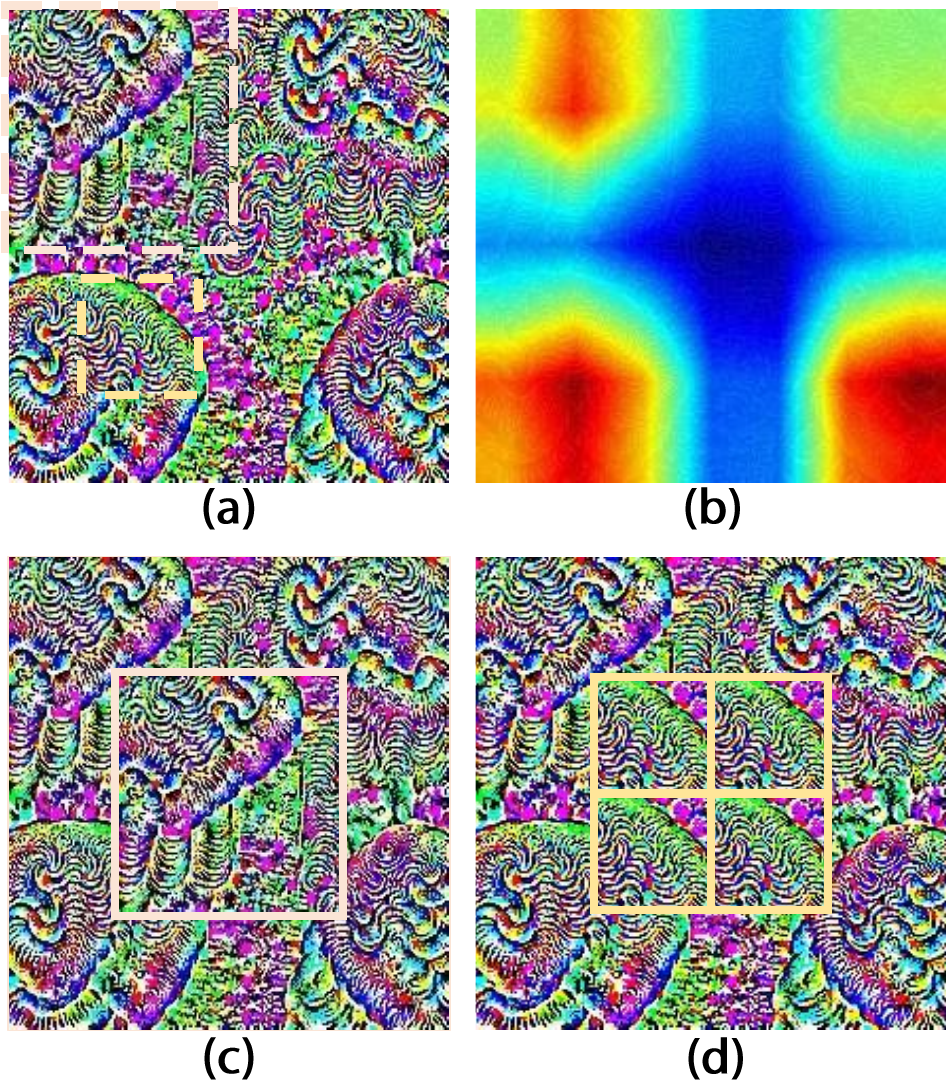}
    \caption{(a): UAP generated by SGD-UAP. The UAP achieves 80.3\% fooling ratio. (b) Attention map of (a). (c): Replace the center of (a) with textures in the top left corner of (a) (\ie, textures in the pink box). The UAP achieves 81.1\% fooling ratio. (d): Replace the center of (a) with the replicated local textures in (a) (\ie, textures in the yellow box). The UAP achieves 84.3\% fooling ratio.}
    \label{fig:small_expr}
\end{figure}
\begin{figure*}[t]
    \centering
    \includegraphics[width=0.75\linewidth]{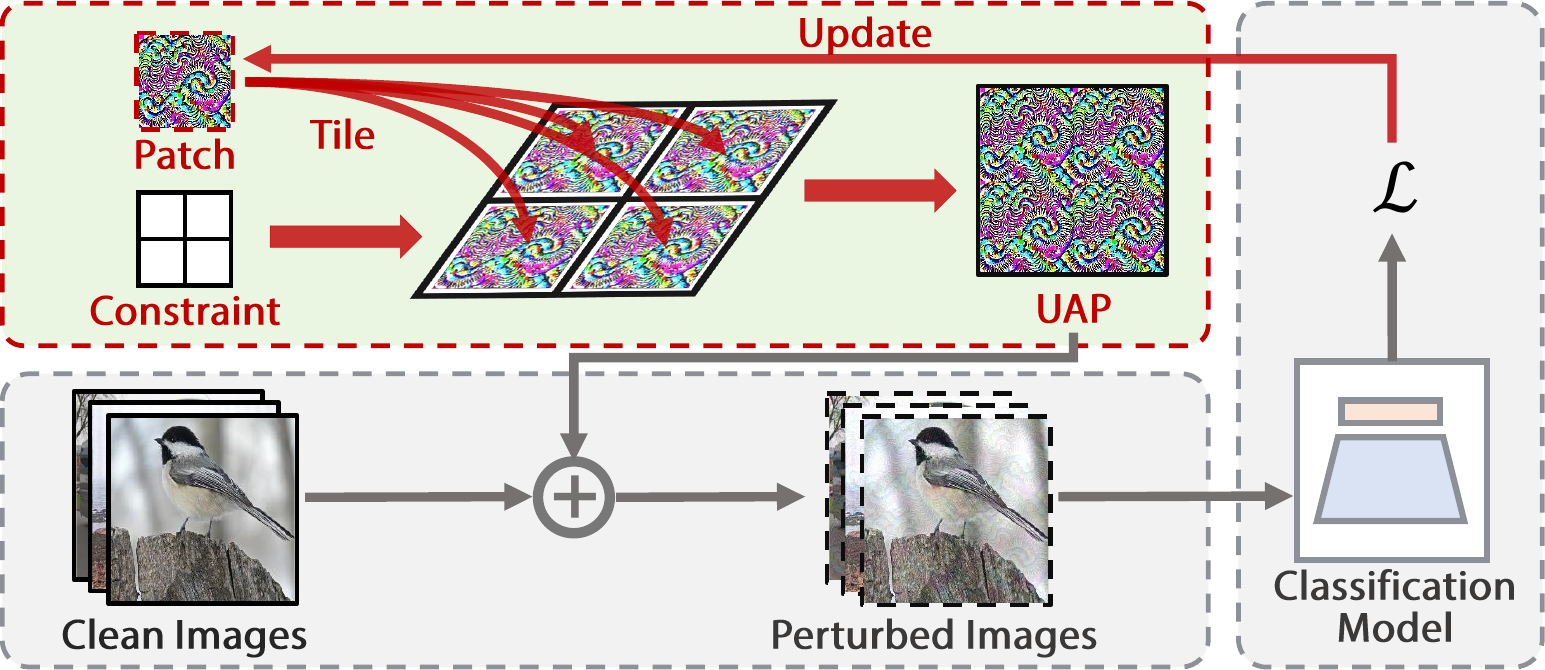}
    \caption{We choose the pipeline of SGD-UAP since the method is concise. It can represent both data-dependent and data-free UAP generation methods. In the standard UAP generation pipeline, they add UAP to the clean images and put the perturbed images into the classifier to calculate the loss. Our modification is in the green box with red bounds. The difference is that we use the patch (\ie, category-specific local texture) and texture scale constraints (\ie, split ratio $\alpha$) to obtain the UAP. The gradient back-propagation is applied on the patch, not the UAP.}
    \label{fig:framework-TSCUAP}
\end{figure*}

\section{Methodology}\label{sec:method}
\subsection{Problem Statement}
Let $\mathcal{X} \in \mathds{R}^{d}$ denotes a distribution of training images and $\hat{\mathcal{C}}(\cdot)$ represents a classifier function. The objective of UAP is to seek a fixed single perturbation vector $\delta \in \mathds{R}^{d}$ which can fool the $\hat{\mathcal{C}}(\cdot)$ on most of the data samples $x \sim \mathcal{X}$. Please note that as a restricted adversarial attack method, $\delta$ should be a tiny noise bounded by $\epsilon$ on the $l_{p}$-norm. Specifically, UAP aims to seek $\delta$ such that 
\begin{align}
 \hat{\mathcal{C}}(x+\delta) \neq \hat{\mathcal{C}}(x)~\text{for~most}~x \sim \mathcal{X}, ~\text{s.t.}~   \|\delta\|_{p} \leq \epsilon.
\label{eq:UAP_def}
\end{align}
In previous works, $\epsilon = 10/255$ and $p$ = $\infty$ is a most common choice for image value ranges in $[0,1]$.

The most commonly used metric to evaluate the effectiveness of UAP is the \textit{fooling ratio}, which is defined as:
\begin{align}
\frac{| \{x \in \mathcal{X}_{test}: \hat{\mathcal{C}}(x+\delta) \neq \hat{\mathcal{C}}(x) \}|}{N},
\label{eq:fooling_ratio_def}
\end{align}
where $N$ is the size of test dataset $\mathcal{X}_{test}$.

\subsection{Motivation}\label{sec:motivation}
Although previous UAP generation methods have achieved a relatively high fooling ratio (around 80-90\%) \cite{ijcai2021p635}, there is still room for improvement.
Here, we use the UAP generated by SGD-UAP \cite{shafahi2020universal} as an example to illustrate our observation. The reason for selecting SGD-UAP is that its objective function is easy to understand, and it is representative since the algorithms of state-of-the-art UAP methods, such as DF-UAP \cite{zhang2020understanding} and Cos-UAP \cite{zhang2021data}, are based on it. The objective function is
\begin{align}
\max_\delta{\mathcal{L} = \frac{1}{N_{\mathcal{X}}}\sum_{i=1}^{N_{\mathcal{X}}} l(x_i+\delta)},~\text{s.t.}~\|\delta\|_{p} \leq \epsilon,
\label{eq:SGD-UAP-obj}
\end{align}
where $l(\cdot)$ is the loss function (different in SGD-UAP, DF-UAP and Cos-UAP) and $N_{\mathcal{X}}$ is the size of dataset $\mathcal{X}$.

As shown in Figure~\ref{fig:small_expr}(a), the UAP is generated by SGD-UAP on ResNet50 with ImageNet training samples and achieves an 80.3\% fooling ratio on the validation set. There are three observations about this UAP. \ding{182} We can observe that the textures in UAP are very similar to the pattern of coral. Such obvious category-related texture can not be observed in non-universal adversarial noise. \ding{183} This UAP, if added to the clean image $x$, will make the ResNet50 misclassify the fusion image as `coral' (\ie, $\mathcal{\hat{C}}$($\delta+x$)=`coral') with a high probability. \ding{184} The UAP itself is also classified as `coral' if directly put UAP into the ResNet50 (\ie, $\mathcal{\hat{C}}$($\delta$)=`coral'). Please note that the observation \ding{183} and \ding{184} are claimed in the paper of DF-UAP and Cos-UAP in a more general form. Due to the category consistency between visualization and the model's prediction, we conclude that the category-related textures highly influence the category of UAP. Since UAP dominates the classification of fusion images, we further conclude that the category-related texture may influence the attack performance of UAP. The conclusions lead us to focus our research on the textures of UAP.

With our naked eye, we can find obvious coral patterns in the four corners of UAP and the textures in the center of UAP are not distinctly coral-like. From the perspective of CNN, it also gives a similar viewpoint (we generate an attention map by Grad-cam \cite{selvaraju2017grad} to show which area CNN models take as the evidence to classify UAP as coral. The result (\ie, Figure~\ref{fig:small_expr}(b)) highlights the four corners). Since the category-related textures may influence the attack performance of UAP, it brings an interesting question: what will happen to the fooling ratio if we replace the insipid (less category-related) textures with the textures of strong category features? Figure~\ref{fig:small_expr}(c) is achieved by cropping the quarter top left corner of UAP (\ie, texture in the pink box of Figure~\ref{fig:small_expr}(a)) and pasting it to the center of the original UAP. This manually constructed UAP is coarse because it simply moves the distinct category-related textures to the center (not a carefully calculated location) and does not consider the coherence of textures. Surprisingly, such a coarse UAP obtains an 81.1\% fooling ratio, which is higher than the original UAP. This observation inspires us to tile category-related textures into the UAP for better performance.

Furthermore, CNN models tend to classify objects according to local textures rather than global shapes \cite{geirhos2018imagenet}, thus maybe pasting relatively small local textures into the UAP will achieve a better fooling ratio. To verify this hypothesis, as shown in Figure~\ref{fig:small_expr}(d), we crop one-sixteenth of UAP (\ie, texture in the yellow box of Figure~\ref{fig:small_expr}(a)) that catches model attention and tile four same textures into the center of original UAP. This new UAP fools the CNN at a ratio of 84.3\%, which is significantly higher than that of the original UAP. This observation inspires us to research the scale of texture to improve the fooling ratio of UAP.

In conclusion, according to the toy examples, \textit{we suggest adding ``category-specific local textures'' into the UAP and this may benefit the attack performance of UAP}. The explanation of why we use the ``category-specific local texture'' instead of ``category-related local texture'' is in Sec.~\ref{sec:explanation_category-specific}. Although we can manually select the category-specific local textures and add them to the UAP, the procedure is subjective because we need to empirically choose local textures. Thus we face a challenge on how to automatically generate UAP including category-specific local textures. To solve the problem, we propose texture scale-constrained UAP.

\subsection{Texture Scale Constrained UAP}\label{sec:TSC-UAP}
To generate UAP with category-specific local textures, an intuitive idea is to segment UAP into many small regions and ensure each region to be a special texture. Since the generation of UAP is based on gradient back-propagation towards full-scale UAP, it requires processing the gradient of UAP in a fine-grained way to fit various regions, which is complex and difficult. 
Thus we propose to think the other way around and simplify the problem, that is, given a category-specific local texture patch $v$, how to construct a bigger patch (\ie, UAP) that has the same size as the training images? It is intuitive that tiling the patch $v$ to be a bigger patch is reasonable for that tile is a regular image processing operation with minor cost. To further simplify the operation, for a given training image $\mathbf{I} \in \mathds{R}^{H \times W \times 3}$, we suggest exploiting a uniform split ratio $\alpha$ which is the common divisor of $H$ and $W$ to segment the UAP  $\delta \in \mathds{R}^{H \times W \times 3}$. Thus the category-specific local texture is of shape $(\frac{H}{\alpha} \times \frac{W}{\alpha} \times 3)$. With the category-specific local texture $v$ and tile function $\mathcal{T}(\cdot)$, we can obtain the UAP with formula
\begin{align}
\delta = \mathcal{T}(v,\alpha).
\label{eq:tile_function}
\end{align}
With the split ratio $\alpha$, we can flexibly control the shape of category-specific local textures, and we named such an operation texture scale constraint (TSC). The objective function of TSC-UAP can be extended from Eq.~\eqref{eq:SGD-UAP-obj} to 
\begin{small}
\begin{align}
\max_v{\mathcal{L} = \frac{1}{N_{\mathcal{X}}}\sum_{i=1}^{N_{\mathcal{X}}} l(x_i+\mathcal{T}(v,\alpha))},~\text{s.t.}~\|\mathcal{T}(v,\alpha)\|_{p} \leq \epsilon.
\label{eq:TSC-UAP-obj}
\end{align}
\end{small}

\begin{algorithm}[tb]
	{
		\caption{TSC-UAP}\label{alg:alg_TSCUAP}
		\KwIn{Training dataset $\mathcal{X}$, Classifier $\hat{\mathcal{C}}$, Batch size $m$, Number of epochs $E$, Patch $v$, Perturbation magnitude $\epsilon$, Split ratio $\alpha$}
		\KwOut{Perturbation $\delta$}
            $v \gets 0$ \Comment*[r]{initialization}
            \For{$epoch = 1\ \mathrm{to}\ {E}$}
            {\label{line:epoch}
                $I \gets |\mathcal{X}|/m$ \label{line:cal_iteration} \Comment*[r]{iteration number}
     		\For{$iteration = 1\ \mathrm{to}\ {I}$}{
                    $\delta = \mathcal{T}(v,\alpha)$ \Comment*[r]{generate UAP}
     		    $B \sim \mathcal{X}: |B|=m$ \Comment*[r]{randomly sample}
     		    $g_{v} \gets \mathop{\mathds{E}}\limits_{x \sim B}[\nabla_{v}\mathcal{L}]$  \Comment*[r]{gradient}
     		$v \gets \mathrm{Optim}(g_{v}$) \Comment*[r]{update patch}
                $v \gets \mathrm{min}(\epsilon,\mathrm{max}(v,-\epsilon))$ \Comment*[r]{clipping}\label{alg:clipping}
                }
            $\delta = \mathcal{T}(v,\alpha)$\ \Comment*[r]{generate UAP}
            }
	}
\end{algorithm}
\begin{figure*}[tb]
    \centering
    \includegraphics[width=0.9\linewidth]{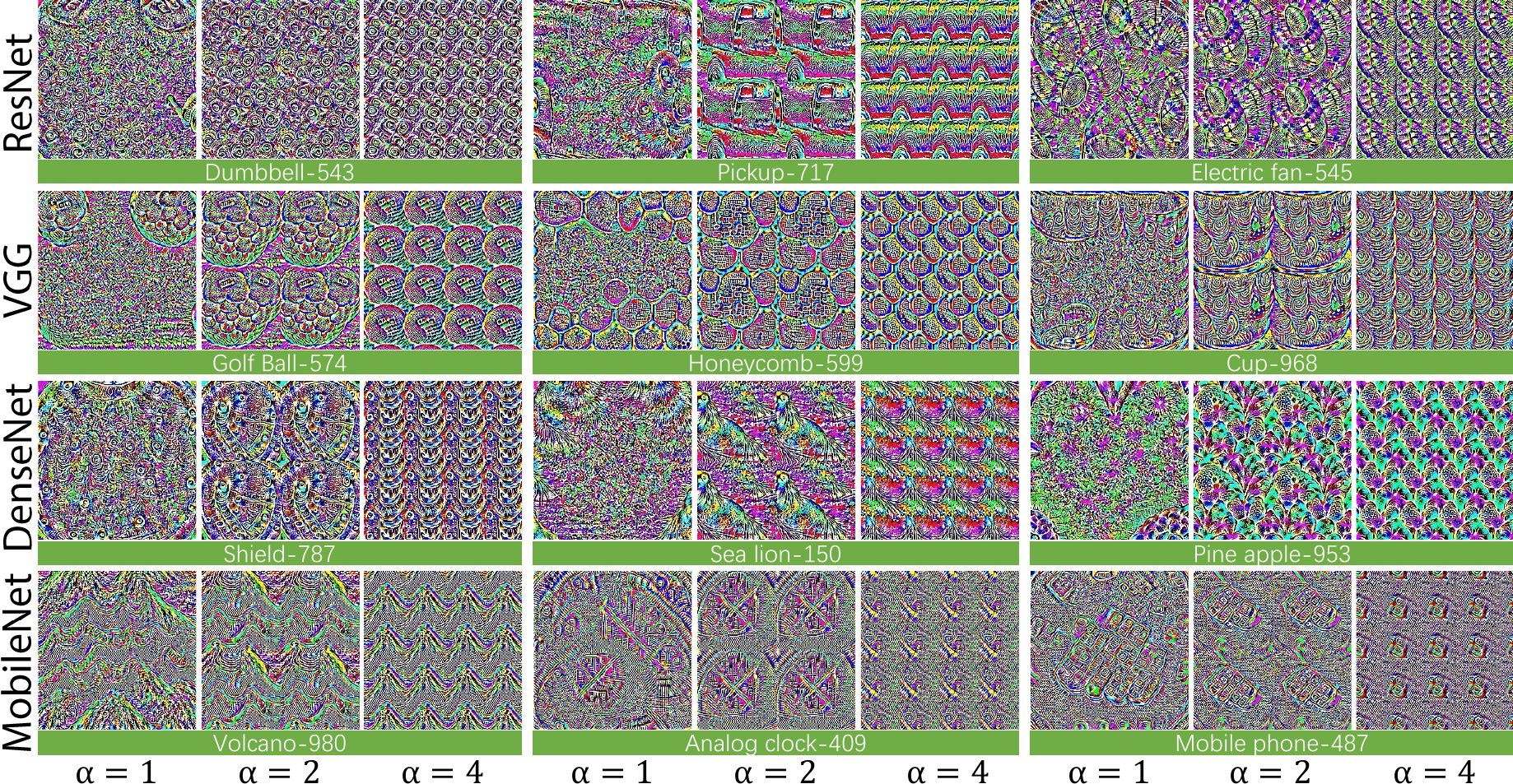}
    \caption{Here shows the human understandable UAPs generated by attacking ResNet50, VGG19, DenseNet121 and MobileNet-v2 on ImageNet training samples.}
    \label{fig:show_human_understandable_UAPs}
\end{figure*}

Referring to the texture scale constraints, we enhance the common pipeline of UAP generation methods. As shown in Figure~\ref{fig:framework-TSCUAP}, the modification is in the green box and the standard UAP generation pipeline is in the gray boxes. Here we choose the pipeline of SGD-UAP since the method is concise and popular. The pipeline can represent both data-dependent and data-free UAP generation methods, which means our texture scale constraints can be applied to both of them, showing the generality of TSC. Furthermore, the state-of-the-art UAP generation methods (DF-UAP \cite{zhang2020understanding} and Cos-UAP \cite{zhang2021data}) share the same pipeline as SGD-UAP (only the loss function is different), which means our texture scale constraints can also improve the SOTA UAP generation methods. To be specific, the loss function of DF-UAP for each data sample is $\mathcal{L} = \max(\mathcal{C}_{gt}(x_i+\delta)-\max_{j \neq gt}\mathcal{C}_j(x_i+\delta), -\kappa)$, where $x_i$ is the sample from the training dataset $\mathcal{X}$, $\mathcal{C}_j$ indicate the $j$-th value of the output logit, $gt$ indicate the ground truth label of sample $x_i$, $\kappa$ is the confidence value set by researcher. The loss function of Cos-UAP for each data sample is $\mathcal{L} = {CosSim}(\mathcal{\hat{C}}(x_i),\mathcal{\hat{C}}(x_i+\delta))$, where $CosSim$ is cosine similarity, $x_i$ is the sample from the training dataset $\mathcal{X}$. In the standard UAP generation pipeline, they add UAP to the clean images and put the perturbed images into the classifier to calculate the loss. Please note that the loss is calculated for each image batch, not individual images. The update of the UAP is based on the gradient calculated by the loss. 
In our method, the modification is that we initialize the patch $v$ and texture scale constraints with the split ratio $\alpha$. Then we tile the patch to obtain the UAP. The gradient back-propagation is applied on the patch, not the UAP, which is computationally efficient. Algorithm~\ref{alg:alg_TSCUAP} outlines the procedure of TSC-UAP. Please note that since $L_\infty$ norm is most commonly used in UAP generation methods, the Algorithm~\ref{alg:alg_TSCUAP} is the implementation of Eq.~\ref{eq:TSC-UAP-obj} in $L_\infty$ norm. Specifically, the constraint $\|\mathcal{T}(v,\alpha)\|_{\inf} \leq \epsilon$ forces the largest absolute value of $\mathcal{T}(v,\alpha)$ to be smaller than $\epsilon$. The Line \ref{alg:clipping} of algorithm is to clip the absolute value within $v$ larger than $\epsilon$, which is equal to $\|\mathcal{T}(v,\alpha)\|_{\inf} \leq \epsilon$ since $\mathcal{T}(v,\alpha)$ is tiled by the patch $v$.

\section{Explanation on ``category-specific local texture''}\label{sec:explanation_category-specific} The ``category-specific local texture'' means that we can generate a specific and local universal adversarial texture for each class category (cat, dog, \etc) and the local textures of different categories are different. Note the `specific' property is not exclusive to our local textures, and existing UAPs (\eg, the UAPs generated by SGD-UAP, Cos-UAP, DF-UAP) are also category-specific. The explanation is as follows.

Given a dataset $\mathcal{X}$ and a targeted category $y_i$, we optimize a targeted UAP $\delta$ by solving the following objective function: 
$\delta^*_i=\text{argmin}_{\delta} \sum_{x\in\mathcal{X}} l(\mathcal{\hat{C}}(x+\delta),y_i),~\text{s.t.}~\|\delta\|_{p} \leq \epsilon$. If the dataset $\mathcal{X}$ is fixed and we use the same optimization algorithm for different categories, we have $\delta^*_i\approx\delta^*_j,~\text{if}~y_i=y_j$ and $\delta^*_i\neq\delta^*_j, \text{if}~y_i\neq y_j$, which actually means the generated UAPs are category specific. The term `$\approx$' is caused by random operations during optimization like example sampling. 
In terms of the untargeted attack, although it does not need the targeted category, there is an intriguing phenomenon of untargeted UAP explained in \cite{zhang2021data}: most images are misclassified to a dominant label, which is the same as the targeted UAP. Hence, untargeted UAP is also category-specific.

The above statement can be extended to our local UAP textures directly. Thus, we say UAPs generated by TSC-UAP contain category-specific local textures.
Please note that although the CNNs can recognize category-specific local textures in UAPs and category-specific local textures can benefit the attack performance of UAPs, the shape of textures should not be too small. In specific, we empirically find that when the split ratio $\alpha>8$ on ImageNet images, the local texture usually may not be human-understandable and fail to enhance the attack performance of UAPs. In order to facilitate the understanding of ``category-specific local texture'', we select human-understandable UAPs of various categories on different CNNs in Figure~\ref{fig:show_human_understandable_UAPs} with $\alpha =  1/2/4$. We can find that UAPs of different categories have distinct appearances and are related to category patterns.
For example, UAPs of the category `Shield' have shield-like textures. Furthermore, as the split ratio $\alpha$ increases, TSC-UAP is more likely to extract the local texture of objects. 

\section{Experiment}\label{sec:experiment}
\noindent\textbf{Datasets and comparative methods.} We evaluate the proposed TSC-UAP on a variety of official pre-trained CNN in PyTorch, including GoogleNet \cite{szegedy2015going}, VGG \cite{simonyan2014very}, ResNet \cite{he2016deep}, DenseNet \cite{huang2017densely}, MobileNet-v2 \cite{sandler2018mobilenetv2}, \etc. We use the ImageNet validation set \cite{russakovsky2015imagenet} (50,000 images) to evaluate the performance of TSC-UAP. If not specified, the test is done on the entire validation set. With regard to the training dataset of UAP generation, if not specified, the size is 1,000 (by sampling one image for each class (1000 classes) in the ImageNet training dataset). We also carry out experiments on CIFAR10 \cite{krizhevsky2009learning}, CIFAR100, and Places-365 \cite{zhou2017places} for further verification. Places-365 is a scene recognition dataset that has 365 classes. We compare the attack performance of TSC-UAP with classical SGD-UAP and SOTA DF-UAP and Cos-UAP. Please note that the DF-UAP and Cos-UAP both do not provide the official code. Since the only difference between them and SGD-UAP is the loss function, we implement these two UAP methods based on the SGD-UAP pipeline and only change the loss function, termed ``DF-UAP-rep'' and ``Cos-UAP-rep'' respectively. 

\noindent\textbf{Evaluation metric.} We use the widely acknowledged fooling ratio metric to quantitatively measure performance.

\noindent\textbf{Implementation details.} The maximum perturbation magnitude is set to 10/255 with the pixel range in [0,1]. The attack epoch is 20 and the batch size is 100. The loss function is cross-entropy. We use $tile(\cdot)$ of PyTorch as the tile function $\mathcal{T}(\cdot)$, which is differentiable. For hyper-parameter $\alpha$, we set it to be one of 1, 2, 4, 8, 16, 32 since they evenly scale the texture size and they are divisible by 224 (the common input shape of CNNs). It is obvious that $\alpha$ = 1 means there are no texture scale constraints on the UAP. Since we actually do not add any TSC constraint on SGD-UAP, it is not suitable to regard such a situation as our TSC-UAP. We use $\alpha=1$ in experiments aiming to make a unified representation ($\alpha=1,2,4,8,16,32$) and facilitate readers to confirm the effects of TSC-UAP under different split ratios. The choice of $\alpha$ values may be an interesting problem, which will be investigated in future work. All the experiments were run on a Ubuntu system with an NVIDIA GeForce RTX 3090 of 24G RAM.
\begin{table}[tb]
\center
\setlength{\tabcolsep}{2.5pt}
\caption{Target models (\ie, ResNet50, VGG19, DenseNet121, MobileNet-v2) are in the first row and the texture scale parameter $\alpha$ is in the first column. In each cell, the highest fooling ratio is in bold and exceeded the baseline by an average of 11.77\%.}
\resizebox{\linewidth}{!}{
\begin{tabular}{c|cccc}
\toprule 
\diagbox{$\alpha$}{Model} & ResNet50 & VGG19 & DenseNet121 & MobileNet-v2\tabularnewline
\hline 
1 & 80.30 & 82.44 & 66.05 & 94.08\tabularnewline
2 & 87.22 & 90.37 & 78.83 & 97.01\tabularnewline
4 & 89.92 & \textbf{92.88} & 82.26 & 97.25\tabularnewline
8 & \textbf{92.73} & 86.57 & \textbf{85.18} & \textbf{99.19}\tabularnewline
16 & 78.47 & 83.54 & 79.26 & 97.53\tabularnewline
32 & 53.96 & 79.39 & 51.24 & 74.65\tabularnewline
\bottomrule 
\end{tabular}}
\label{tab:fr-TSC-SGDUAP}
\end{table}
\begin{figure}[tbp]
    \centering
    \includegraphics[width=\linewidth]{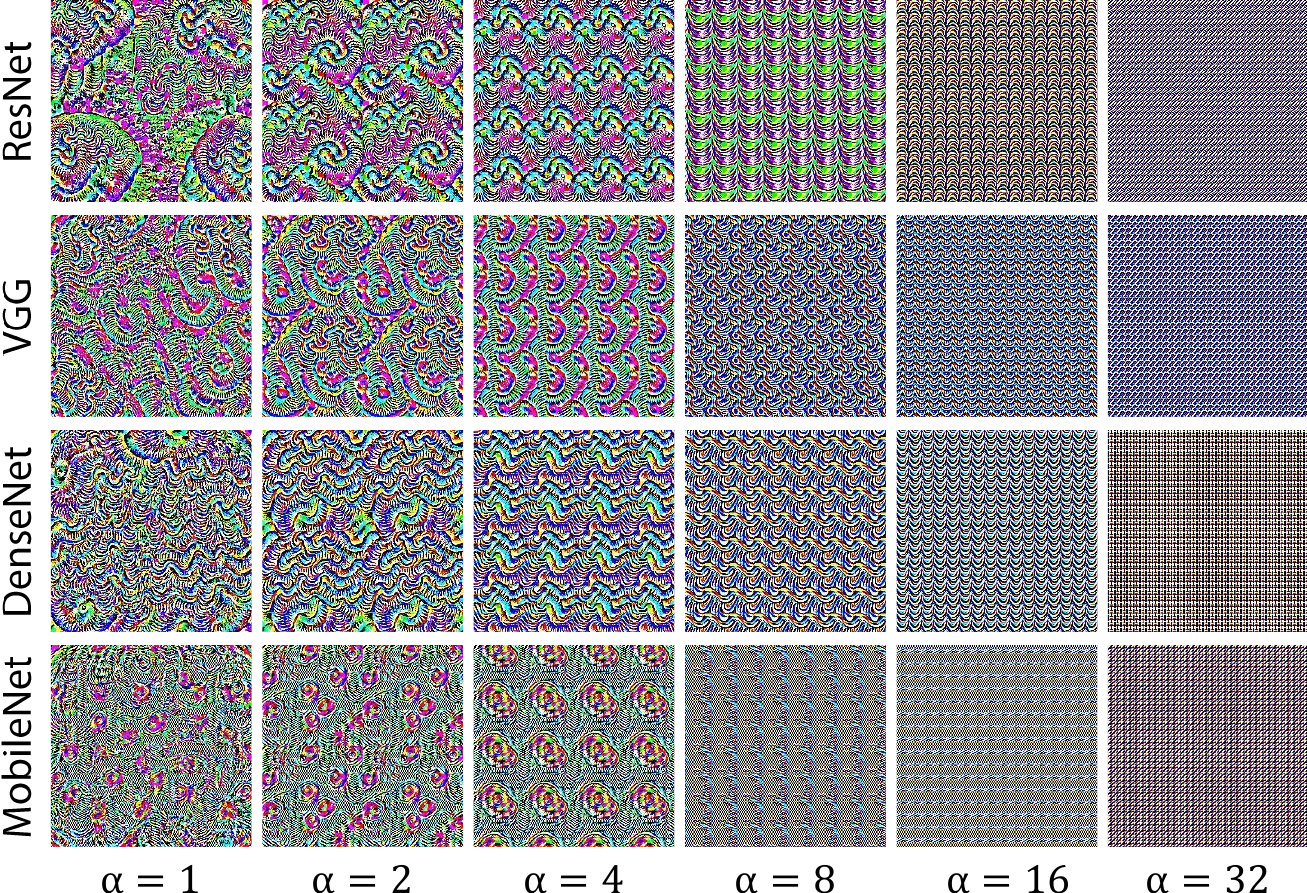}
    \caption{Untargeted UAPs generated with different texture scale constraints (\ie, $\alpha$) on ImageNet training samples.}
    \label{fig:show_UAPs}
\end{figure}

\begin{table*}[tb]
\center
\setlength{\tabcolsep}{2.5pt}
\caption{First two columns represent the UAPs generated by source models, which is the same as that in Table~\ref{tab:fr-TSC-SGDUAP}. First two rows represent the target models which need to attack. In each cell, there shows six fooling ratio values, corresponding to six different texture scales (\ie, $\alpha$ = 1/2/4/8/16/32). The highest fooling ratios are in red and exceeded the baseline by an average of 21.44\%.}
\resizebox{\linewidth}{!}{
\begin{tabular}{c|l|cccc}
\toprule 
\multicolumn{2}{c|}{\multirow{2}{*}{FR($\alpha$ = 1/2/4/8/16/32)(\%)}} & \multicolumn{4}{c}{Target}\tabularnewline
\cline{3-6} 
\multicolumn{2}{c|}{} & ResNet50 & VGG19 & DenseNet121 & MobileNet-v2\tabularnewline
\hline 
\multirow{4}{*}{Source} & ResNet50 & - & 53.03/59.45/63.83/\textcolor{red}{80.13}/61.28/73.13 & 41.64/49.91/58.12/\textcolor{red}{66.16}/51.90/33.97 & 51.50/57.12/63.99/\textcolor{red}{72.28}/55.93/63.98\tabularnewline
 & VGG19 & 32.64/43.88/48.61/\textcolor{red}{48.73}/42.75/34.61 & - & 29.18/38.73/43.26/\textcolor{red}{43.99}/36.02/28.44 & 45.95/56.72/\textcolor{red}{63.71}/56.40/48.32/55.04\tabularnewline
 & DenseNet121 & 42.89/53.29/57.48/60.13/\textcolor{red}{71.58}/49.05 & 50.74/58.44/64.09/75.18/\textcolor{red}{79.54}/64.18 & - & 49.25/54.39/59.18/65.66/66.05/\textcolor{red}{67.78}\tabularnewline
 & MobileNet-v2 & 26.95/35.32/39.21/34.75/32.89/\textcolor{red}{42.47} & 38.18/47.85/50.76/41.17/36.01/\textcolor{red}{66.90} & 23.23/29.51/31.83/22.51/21.47/\textcolor{red}{39.28} & -\tabularnewline
\bottomrule 
\end{tabular}}
\label{tab:fr-TSC-transfer}
\end{table*}

\subsection{Improve Fooling Ratio}\label{sec:untarget_basic_attack}
\begin{table*}[htb]
\center
\caption{We randomly select 5 target attack labels. The first and third rows represent the target models. In each cell of the first/last two rows, there are six fooling ratios/target fooling ratios, corresponding to six different texture scales $\alpha$ = 1/2/4/8/16/32. The highest fooling ratios/target fooling ratios are in red and exceeded the baseline by an average of 20.68\% and 20.3\% respectively.}
\resizebox{\linewidth}{!}{
\begin{tabular}{l|cccc}
\toprule 
FR($\alpha$ = 1/2/4/8/16/32)(\%)& ResNet50 & VGG19 & DenseNet121 & MobileNet-v2\tabularnewline
\hline 
echidna-102 & 22.23/\textcolor{red}{58.75}/47.75/55.99/41.10/37.60 & 63.45/\textcolor{red}{78.78}/72.55/66.46/56.37/70.88 & 43.50/\textcolor{red}{65.06}/61.78/44.55/39.10/32.05 & 54.14/64.33/\textcolor{red}{67.47}/56.78/62.82/57.91\tabularnewline
starfish-327 & 46.63/\textcolor{red}{69.52}/64.35/61.43/65.66/40.50 & 67.40/82.71/\textcolor{red}{83.24}/73.14/61.40/63.95 & 49.13/61.91/\textcolor{red}{66.28}/55.79/49.99/42.38 & 74.55/\textcolor{red}{78.60}/69.12/76.63/63.68/62.92\tabularnewline
golf ball-574 & 44.69/65.98/\textcolor{red}{72.00}/64.39/44.63/43.17 & 58.51/72.18/\textcolor{red}{78.33}/66.91/74.97/65.71 & 26.92/54.56/64.92/\textcolor{red}{68.20}/51.46/34.67 & 82.56/85.13/85.49/\textcolor{red}{88.88}/78.22/67.60\tabularnewline
parachute-701 & 29.25/64.13/\textcolor{red}{70.24}/50.11/37.88/40.61 & 63.49/\textcolor{red}{81.92}/75.74/71.44/61.72/55.87 & 43.98/\textcolor{red}{63.13}/62.92/52.44/35.53/35.82 & 66.62/\textcolor{red}{74.03}/70.07/64.88/68.60/64.29\tabularnewline
pineapple-953 & 51.12/\textcolor{red}{74.07}/73.83/69.88/65.83/45.07 & 52.15/\textcolor{red}{80.03}/79.48/73.83/67.94/65.50 & 35.35/60.27/\textcolor{red}{66.47}/60.75/49.15/42.81 & 87.85/91.29/\textcolor{red}{92.13}/91.59/76.98/68.01\tabularnewline
\toprule 
TFR($\alpha$ = 1/2/4/8/16/32)(\%)& ResNet50 & VGG19 & DenseNet121 & MobileNet-v2\tabularnewline
\hline 
echidna-102 & 0.10/\textcolor{red}{30.96}/2.82/0.15/0.13/0.10 & 32.91/\textcolor{red}{50.20}/38.63/7.07/0.29/0.08 & 17.47/\textcolor{red}{43.09}/20.27/3.55/0.13/0.22 & 8.08/\textcolor{red}{17.90}/2.31/0.35/0.12/0.12\tabularnewline
starfish-327 & 22.38/\textcolor{red}{37.52}/13.80/6.06/1.13/0.18 & 38.96/\textcolor{red}{55.68}/52.00/14.86/4.26/1.36 & 29.54/36.43/\textcolor{red}{49.40}/18.36/0.75/0.15 & 36.50/\textcolor{red}{47.77}/27.18/7.69/3.55/1.17\tabularnewline
golf ball-574 & 14.94/27.28/\textcolor{red}{39.65}/24.57/0.04/0.03 & 29.07/\textcolor{red}{46.97}/43.07/0.27/0.23/0.28 & 7.19/34.99/\textcolor{red}{44.33}/41.86/3.06/0.15 & 33.74/\textcolor{red}{35.63}/6.83/1.05/0.28/0.11\tabularnewline
parachute-701 & 5.42/40.92/\textcolor{red}{43.73}/2.25/0.21/0.16 & 33.52/\textcolor{red}{53.23}/41.43/17.45/1.61/0.15 & 22.25/\textcolor{red}{39.36}/37.08/9.49/0.21/0.28 & 18.10/\textcolor{red}{26.40}/8.41/2.82/0.65/0.41\tabularnewline
pineapple-953 & 21.11/\textcolor{red}{42.77}/35.12/31.03/1.19/0.01 & 6.13/\textcolor{red}{35.69}/29.57/0.80/0.05/0.01 & 12.81/38.33/\textcolor{red}{43.14}/23.73/1.00/0.01 & 47.82/\textcolor{red}{60.65}/40.99/5.38/0.16/0.02\tabularnewline
\hline 
\end{tabular}}
\label{tab:fr-TSC-target}
\end{table*}
\begin{table*}[htb]
\center
\caption{Target models are in the first row and the datasets are in the first column. In each cell, there are six fooling ratios, corresponding to six different texture scales $\alpha$ = 1/2/4/8/16/32. For CIFAR10, CIFAR100 and Place365, the highest fooling ratio exceeded the baseline by an average of 15.32\%, 25.37\%, and 17.93\% respectively.}
\resizebox{\linewidth}{!}{
\begin{tabular}{l|cccc}
\toprule 
FR($\alpha$ = 1/2/4/8/16/32)(\%) & ResNet50 & VGG19 & DenseNet121 & MobileNet-v2\tabularnewline
\hline 
CIFAR10 & 46.52/66.93/75.01/\textcolor{red}{88.92}/64.11/6.05 & 89.33/89.64/\textcolor{red}{89.68}/89.04/87.69/88.04 & 69.43/81.86/84.74/\textcolor{red}{86.41}/84.01/6.55 & 87.58/\textcolor{red}{89.16}/88.21/85.38/85.14/82.72\tabularnewline
CIFAR100 & 37.81/55.15/\textcolor{red}{62.51}/52.41/38.96/8.85 & 71.92/85.22/\textcolor{red}{89.93}/79.13/81.08/77.80 & 63.20/78.62/\textcolor{red}{87.99}/84.29/63.03/13.13 & 89.34/90.17/92.52/85.45/84.75/\textcolor{red}{94.21}\tabularnewline
Place365 & 39.14/41.83/\textcolor{red}{45.87}/40.12/40.99/30.37 & - & - & 35.81/37.99/37.96/21.30/20.90/\textcolor{red}{64.94} \tabularnewline
\bottomrule 
\end{tabular}}
\label{tab:fr-TSC-cross-data}
\end{table*}

We first show the improvement of the fooling ratio on the standard data-dependent UAP generation method (\ie, SGD-UAP). The experiment on SGD-UAP is able to verify the effectiveness of the texture scale constraints on a common UAP pipeline, which shows the wide application range of texture scale constraints.

In Table~\ref{tab:fr-TSC-SGDUAP}, target models (\ie, ResNet50, VGG19, DenseNet121, MobileNet-v2) are in the first row and choices of $\alpha$ are in the first column. For each target model, the highest fooling ratio is in bold. For example, in the second column (`ResNet50'), the performance of baseline is 80.3\% (\ie, when $\alpha$ = 1). The highest fooling ratio is 92.73\%, which is more than 10\% higher than the baseline and achieved when $\alpha$ is 8. With respect to VGG, DensetNet, and MobileNet, the highest fooling ratio all outstrip the baseline and the improvements are 10.44\%, 19.13\%, and 5.11\% respectively. Even though the fooling ratio of the baseline is high (\ie, more than 90\% in `MobileNet'), texture scale constraints can still enhance the performance of UAP. Among all four models, when $\alpha \leq 8$ and in the vast majority of cases of $\alpha$ = 16, the fooling ratios are higher than the baseline, showing that texture scale constraints can stably improve the UAP.

Furthermore, we find a regular phenomenon that the fooling ratios usually first rise then descend with the increasing of $\alpha$ and peak around $\alpha$ = 8. We think this mainly because 
%
%
an overly tiny texture scale means that only a few parameters can be optimized with the gradient. Restricted by perturbation magnitude $\epsilon$, this is a really small search space, which is not enough for generating UAP with a high fooling ratio. Thus we empirically suggest setting $\alpha$ around 8. 

The generated UAPs are shown in Figure~\ref{fig:show_UAPs}. We can find that with the increase of $\alpha$, the UAPs include small-scale category-specific textures.

\begin{table*}[tb]
\centering
\caption{The TSC-UAP without texture scale constraints (\ie, $\alpha$ = 1) needs 3789.236s to perform the attack while other TSC-UAP with $\alpha$ from 2 to 32 need similar time as it. The running time is even shorter than baseline when $\alpha$ is 8, 16, 32.}
\begin{tabular}{c|cccccc||cc}
\hline 
\multirow{2}{*}{time(s)} & \multicolumn{6}{c||}{TSC-UAP} & \multirow{2}{*}{DF-UAP-rep} & \multirow{2}{*}{Cos-UAP-rep}\tabularnewline
\cline{2-7}
 & $\alpha$ = 1 & $\alpha$ = 2 & $\alpha$ = 4 & $\alpha$ = 8 & $\alpha$ = 16 & $\alpha$ = 32 &  & \tabularnewline
\hline 
ResNet & 3789.236 & 3793.574 & 3795.700 & 3765.133 & 3776.235 & 3769.121 & 3792.637 & 3782.797\tabularnewline
\hline 
\end{tabular}
\label{tab:fr-TSC-time}
\end{table*}
\begin{table*}[tb]
\center
\setlength{\tabcolsep}{2.5pt}
\caption{We compare TSC-UAP with SOTA UAP methods on other datasets. TSC-UAP achieves the best fooling ratio.}
\resizebox{\linewidth}{!}{
\begin{tabular}{l|ccc|ccc|ccc|ccc}
\toprule 
\multirow{2}{*}{FR(\%)} & \multicolumn{3}{c|}{ResNet50} & \multicolumn{3}{c|}{VGG19} & \multicolumn{3}{c|}{DenseNet121} & \multicolumn{3}{c}{MobileNet-v2}\tabularnewline
\cline{2-13}
 & TSC-UAP & DF-UAP-rep & Cos-UAP-rep & TSC-UAP & DF-UAP-rep & Cos-UAP-rep & TSC-UAP & DF-UAP-rep & Cos-UAP-rep & TSC-UAP & DF-UAP-rep & Cos-UAP-rep\tabularnewline
\hline 
CIFAR10 & \textbf{89.76} & 89.72 & 81.91 & \textbf{90.75} & 90.73 & 88.56 & \textbf{89.78} & 89.74 & 87.06 & \textbf{90.10} & 90.05 & 88.05\tabularnewline
CIFAR100 & \textbf{98.41} & 98.04 & 95.51 & \textbf{98.84} & 98.40 & 96.89 & \textbf{98.92} & 98.90 & 97.92 & \textbf{98.75} & 98.45 & 97.98\tabularnewline
\bottomrule 
\end{tabular}}
\label{tab:fr-TSC-other-dataset}
\end{table*}
\begin{table*}[tb]
\center
\caption{The increment of TSC on improving the SOTA data-free UAP method Cos-UAP, at an average of 5.2\%.}
\begin{tabular}{l|cccccccc}
\toprule 
FR(\%) & AlexNet & GoogleNet & VGG16 & VGG19 & ResNet50 & ResNet152 & DenseNet121 & MobileNet-v2\\\hline
Cos-UAP-rep & 96.58 & 75.51 & 92.51 & 85.52 & 87.75 & 84.33 & 76.33 & 95.38\\
TSC-Cos-UAP-rep & \textbf{96.86} & \textbf{90.95} & \textbf{92.56} & \textbf{92.09} & \textbf{91.02} & \textbf{88.68} & \textbf{84.55} & \textbf{98.91}\\
\bottomrule 
\end{tabular}
\label{tab:fr-TSC-data-free}
\end{table*}

\subsection{Improve Cross-model Attack Transferability}
Attack transferability is also an important metric for adversarial attacks. Due to the fact that CNN models pay more attention to local textures \cite{geirhos2018imagenet}, we think the UAPs with relatively small texture scales may have better attack transferability. The results are in Table~\ref{tab:fr-TSC-transfer}. Please note that the values on the diagonal of the table are white-box testing results and are already shown in Table~\ref{tab:fr-TSC-SGDUAP}.

The first two columns represent the UAPs generated by source models, which is the same as that in Table~\ref{tab:fr-TSC-SGDUAP}. The first two rows represent the target models which need to attack. In each cell, there shows six fooling ratio (FR) values, corresponding to six different texture scales (\ie, $\alpha$ = 1/2/4/8/16/32). For example, using the ResNet as the source model and VGG as the target model, the fooling ratios of six different UAPs are 53.03\%/59.45\%/63.83\%/80.13\%/61.28\%/73.13\%. We can find that the highest fooling ratio is achieved with $\alpha$ = 8 and is 27.1\% higher than the baseline (\ie, 53.03\%), which is a significant increment. We use red color to mark the highest fooling ratio in each cell. In all the cells, the UAPs generated by TSC-UAP achieve the highest attack transferability, 28.8\% increment at most and 21.44\% on average. We also find the highest fooling ratios occur most when $\alpha$ = 8, achieving 5 in 12 (\ie, 41.6\%). Please note that in the vast majority of cases, using texture scale constraints can achieve a higher fooling ratio, which shows that it can stably improve cross-model attack transferability.

\subsection{Improve Targeted Attack}
For targeted attacks, we conduct experiments on 5 randomly chosen target labels and for each label, the number of training images is 1,000. We apply two metrics (fooling ratio (FR) and target fooling ratio (TFR)). Compared with FR (as formulated in Eq.~\ref{eq:fooling_ratio_def}), TFR additionally requires the prediction label to be the same as the target label. That is,
\small
\begin{align}
\frac{| \{x \in \mathcal{X}_{test}: \hat{\mathcal{C}}(x+\delta) \neq \hat{\mathcal{C}}(x)~\&~\hat{\mathcal{C}}(x+\delta)=y_{target}\}|}{N}.
\label{eq:target_fooling_ratio_def}
\end{align}
The reason for using FR is to evaluate the attack performance of targeted UAP as a ``universal'' perturbation. The reason for using TFR is to further evaluate the attack performance of targeted UAP as a ``target'' perturbation. 

As shown in Table~\ref{tab:fr-TSC-target}, the first column represents the attack labels. The first and third rows represent the target models. In each cell of the first two rows, there are six fooling ratios, corresponding to six different texture scales $\alpha$ = 1/2/4/8/16/32. The highest FRs are in red and exceeded the baseline by an average of 20.68\%. Referring to the TFR evaluation, in each cell of the last two rows, there are also six target fooling ratios, corresponding to six different texture scales $\alpha$ = 1/2/4/8/16/32. The highest TFRs are in red and exceeded the baseline by an average of 20.3\%.

We can find that in all the cells of no matter FR or TFR evaluation, the highest fooling ratio is not achieved by the baseline, which means texture scale constraints also show satisfying effects on improving the targeted UAP task. Please note that in the vast majority of cases when $\alpha \leq 8$, texture scale constraints can stably improve the fooling ratio on targeted attacks, which is the same as on untargeted attacks.

\subsection{Improve Cross-data Attack Transferability}
UAP is also capable of conducting cross-data attacking \cite{liu2019universal}. We conduct an additional experiment on three different classification datasets (\ie, CIFAR10, CIFAR 100, and Place365) to see whether texture scale constraints improve the cross-data transferability. The UAPs are generated by attacking ImageNet towards four models (\ie, ResNet50, VGG19, DenseNet121, MobileNet-v2) and testing on corresponding models in CIFAR10, CIFAR100, and Place365. Place365 only provides ResNet and MobileNet pre-trained models, thus we only show results on attacking them. Since the shape of the image in ImageNet is not the same as in CIFAR10, CIFAR100, and Place365, thus we generate UAP with ImageNet and resize UAP to the target shape.

As shown in Table~\ref{tab:fr-TSC-cross-data}, the target models are in the first row and the datasets are in the first column. In each cell, the highest fooling ratio is in red. We can find that the UAPs generated by using the texture scale constraints achieve the highest fooling ratios in all the cells. The results show the effect of TSC on improving the cross-data attack performance of UAP. For CIFAR10, CIFAR100 and Place365, the average increments are 15.32\%, 25.37\% and 17.93\%. Please note that in most cases, using texture scale constraints can achieve a higher fooling ratio, which shows that texture scale constraints can stably improve cross-data attack transferability.

\subsection{Time Comparison}
To show the influence of texture scale constraints on the efficiency of the method, we evaluate the run time with different $\alpha$.
In Table~\ref{tab:fr-TSC-time}, the UAP generation method without texture scale constraints (\ie, $\alpha$ = 1) needs 3789.236 seconds to perform the attack while other TSC-UAP with $\alpha$ from 2 to 32 need similar time as it. The running time is even shorter when $\alpha$ is 8, 16, 32. The experiment results show that texture scale constraint is a tiny-cost UAP enhancement approach. We also compare the time with state-of-the-art UAP methods. Since DF-UAP and Cos-UAP do not provide the official code, we realize them based on SGD-UAP by simply adapting their special loss functions, termed DF-UAP-rep and Cos-UAP-rep respectively. We can find that the time of TSC-UAP is similar to DF-UAP-rep and Cos-UAP-rep.

\subsection{Performance on Other Datasets}\label{sec:expr_other_dataset}
As shown in Table~\ref{tab:fr-TSC-other-dataset}, we apply TSC-UAP on other datasets (\ie, CIFAR10 and CIFAR100) and compare with state-of-the-art UAP methods on four classifiers (\ie, ResNet50, VGG19, DenseNet121 and MobileNet-v2). Each cell shows the FR result. When TSC-UAP achieves the best fooling ratio, the values of $\alpha$ parameter are 8/8/8/2 in the CIFAR10 row and 4/4/4/4 in the CIFAR100 row for the ResNet50/VGG19/DenseNet121/MobileNet-v2. We can find that the $\alpha$ for achieving the best fooling ratio for each dataset is similar. On each model and dataset, we bold the highest fooling ratios and TSC-UAP always achieves the best fooling ratio.

\subsection{Improve Data-free UAP Methods}\label{sec:expr_data_free_compare}
The UAP methods can be classified into two types: data-dependent and data-free, according to whether need training samples. We have conducted experiments on the data-dependent UAP method. For data-free UAP methods, we choose the Cos-UAP as the baseline. To our best knowledge, it is the best data-free UAP method. We add texture scale constraints on Cos-UAP-rep to get TSC-Cos-UAP-rep.
In Table~\ref{tab:fr-TSC-data-free}, we conduct experiments on eight different classifiers. In each cell, there is the FR result and we bold the higher fooling ratios. We can find that the fooling ratios of TSC-Cos-UAP-rep are higher than that of Cos-UAP-rep in all the cases. The experiment shows TSC can also improve the data-free UAP methods, even the state-of-the-art ones. When TSC-Cos-UAP-rep achieves the best fooling ratio, the values of $\alpha$ parameter are 2/8/2/2/4/2/4/8 for the AlexNet/GoogleNet/VGG16/VGG19/ ResNet50/ResNet152/DenseNet121/MobileNet-v2. We can find that the $\alpha$ for achieving the best fooling ratio is among (2, 4, 8). Note that even if not achieving the best fooling ratio, the fooling ratios of $\alpha$ in (2, 4, 8) are still high. For example, in the GoogleNet column, when $\alpha$ is 2, 4, 8, the fooling ratios of TSC-Cos-UAP-rep are 86.31\%/85.35\%/90.95\%. Although not achieving the best fooling ratio when $\alpha$ is 2 and 4, but the fooling ratios 86.31\% and 85.35\% are still much better than the fooling ratio (75.51\%) of Cos-UAP-rep.

\subsection{Improve L2-norm UAP}\label{sec:L2_norm_expr}
Although existing UAP generation methods all follow the $L_\infty$-norm, it would be interesting to explore whether our method can extend to other $L_p$-norm. Here we use $L_2$-norm as an example to show the result. As shown in Table~\ref{tab:fr-TSC-SGDUAP-L2}, we apply TSC-UAP towards four different target models (\ie, ResNet50, VGG19, DenseNet121, MobileNet-v2) with six different texture scales (\ie, $\alpha$ = 1/2/4/8/16/32) under $L_2$-norm with $\epsilon$ = 40. In each cell, there is the FR result and we bold the highest fooling ratios for each classifier. We can find that the TSC can improve the attack performance of the UAPs by an average of 24.83\%, which is similar to the improvement under $L_\infty$-norm. It is also interesting to see that most of the optimal value of $\alpha$ is 16, which is different from that under $L_\infty$-norm. This observation not only shows different properties of the UAP under various $L_p$-norm bounds but also further verifies the effectiveness and generality of TSC.

\subsection{Compare to SOTA UAP Methods}\label{sec:expr_SOTA_compare}
DF-UAP \cite{zhang2020understanding}, Cos-UAP \cite{zhang2021data}, TRM-UAP \cite{liu2023trm}, and AT-UAP \cite{li2022learning} are the state-of-the-art UAP methods. To make a fair comparison, we refer to the size of their training dataset. That is, the training dataset is of size 10,000 (by sampling 10 images for each class in the ImageNet training dataset). The comparison is on the ImageNet validation set towards five different models they used (\ie, AlexNet, GoogleNet, VGG16, VGG19, ResNet152).

As shown in Table~\ref{tab:fr-TSC-SOTA}, we bold the highest fooling ratios. We can find that all the UAP methods achieve significantly high fooling ratios, almost achieving more than 90\% on all the models. Among them, TSC-UAP obtains the best fooling ratio on every model. When TSC-UAP achieves the best fooling ratio, the value of $\alpha$ parameter are 4/8/4/4/4 for the AlexNet/GoogleNet/VGG16/VGG19/ResNet152. We can find that the $\alpha$ for achieving the best fooling ratio is similar.


\begin{table}[t]
\center
\setlength{\tabcolsep}{2.5pt}
\caption{The performance of TSC under L2-bounded attack.}
\resizebox{\linewidth}{!}{
\begin{tabular}{c|cccc}
\toprule 
\diagbox{$\alpha$}{Model} & ResNet50 & VGG19 & DenseNet121 & MobileNet-v2\tabularnewline
\hline 
1 & 66.49 & 74.99 & 52.38 & 86.92 \tabularnewline
2 & 81.41 & 94.25 & 60.59 & 95.49 \tabularnewline
4 & 91.82 & 94.77 & 86.10 & 99.04\tabularnewline
8 & 91.96 & 96.33 & 86.88 & 99.16\tabularnewline
16 & \textbf{93.33} & \textbf{97.74} & 84.71 & \textbf{99.61} \tabularnewline
32 & 90.66 & 93.60 & \textbf{89.42} & 95.51\tabularnewline
\bottomrule 
\end{tabular}}
\label{tab:fr-TSC-SGDUAP-L2}
\end{table}
\begin{table}[t]
\center
\setlength{\tabcolsep}{2.5pt}
\caption{We compare TSC-UAP with SOTA UAP methods. Our method achieves better attack performance than DF-UAP, Cos-UAP, TRM-UAP and AT-UAP.}
\resizebox{\linewidth}{!}{
\begin{tabular}{l|ccccc|c}
\toprule 
FR(\%) & AlexNet & GoogleNet & VGG16 & VGG19 & ResNet152 & Average \tabularnewline
\hline 
TSC-UAP & 96.78 & \textbf{91.17} & \textbf{97.64} & 97.51 & \textbf{92.70} & \textbf{95.16}\tabularnewline
DF-UAP & 96.17  & 88.94  & 94.30  & 94.98  & 90.08 & 92.89\tabularnewline
Cos-UAP & 96.50 & 90.50 & 97.40  & 96.40 & 90.20 & 94.20\tabularnewline
TRM-UAP & 93.53 & 85.32 & 94.30  & 91.35 & 67.46 & 86.39\tabularnewline
AT-UAP & \textbf{97.01} & 90.82 & 97.51 & \textbf{97.56} & 91.52 & 94.88\tabularnewline
\bottomrule 
\end{tabular}}
\label{tab:fr-TSC-SOTA}
\end{table}
\begin{table*}[tb]
\center
\caption{The first column means the texture scale parameter $\alpha$. The first row means the size of the total training samples. The second row means the classes and number-per-class chosen by us, recorded briefly in format (c,n).}
\begin{tabular}{c|c|cc|ccc|cccc}
\toprule 
\multirow{2}{*}{FR(\%)} & 1 & \multicolumn{2}{c|}{10} & \multicolumn{3}{c|}{100} & \multicolumn{4}{c}{1000}\tabularnewline
\cline{2-11}
 & (1,1) & (1,10) & (10,1) & (1,100) & (10,10) & (100,1) & (1,1000) & (10,100) & (100,10) & (1000,1)\tabularnewline
\hline 
1 & 19.85 & 21.44 & 22.03 & 22.90 & 25.77 & 23.21 & 78.98 & 75.00 & 76.60 & 80.30\tabularnewline
\hline 
2 & 19.30 & 22.02 & 23.33 & 28.24 & 70.56 & 26.45 & 86.70 & 87.30 & 88.07 & 87.22\tabularnewline
\hline 
4 & 21.60 & 24.77 & 41.80 & 74.41 & 78.81 & 82.11 & 90.52 & 91.65 & 90.22 & 89.92\tabularnewline
\hline 
8 & 23.83 & 28.82 & 71.72 & 87.71 & 87.78 & 80.01 & 93.36 & 93.28 & 93.85 & 92.73\tabularnewline
\hline 
16 & 28.25 & 55.93 & 37.56 & 81.33 & 75.61 & 75.05 & 80.26 & 81.25 & 73.58 & 78.47\tabularnewline
\hline 
32 & 31.60 & 56.18 & 43.03 & 46.48 & 44.28 & 44.28 & 52.11 & 46.49 & 52.54 & 53.96\tabularnewline
\bottomrule 
\end{tabular}
\label{tab:fr-TSC-data-efficiency}
\end{table*}

\subsection{Data-efficiency of TSC-UAP}\label{sec:expr_data_efficiency}
The standard UAP generation methods collect gradient information from a large number of training samples to update the UAP pattern. Since we reduce the update area, similar to sparse representation, there may need fewer training samples to generate UAPs. To verify this, we conduct an experiment to evaluate the data efficiency of the UAP method enhanced by texture scale constraints. Please note that for data-free UAP methods, there is no space for improvement, thus we only show the advance in data-dependent UAP methods.

As shown in Table~\ref{tab:fr-TSC-data-efficiency}, we use ResNet 50 as the target model and the first column means the texture scale. The first row means the size of the total training samples. The second row means the classes and number-per-class chosen by us, recorded briefly in format (c,n). For example, when the total number of training samples is 100, we simply exploit three ways to choose the samples. First, we can choose 1 class (\ie, c=1) with 100 samples (\ie, n=100), expressed as (1,100). Second, we can choose 10 classes (\ie, c=10) with 10 samples each (\ie, n=10), expressed as (10,10). Third, we can choose 100 classes (\ie, c=100) with 1 sample each (\ie, n=1), expressed as (100,1). 

We can find that the fooling ratios achieve around 80\% when the size of the training set is 1,000. However, by using the texture scale constraints, at a tenth of the size we can achieve at most 87.78\% fooling ratio. Furthermore, with only ten images, we can improve the fooling ratio of the standard UAP method from 22.03\% to 71.72\%, which is close to the result of using 1,000 training images with the standard UAP method. The experiment results show the data efficiency of using the texture scale constraints.

\subsection{Visualization of Perturbed Images}
\begin{figure*}[tbp]
\centering
\includegraphics[width=0.8\linewidth]{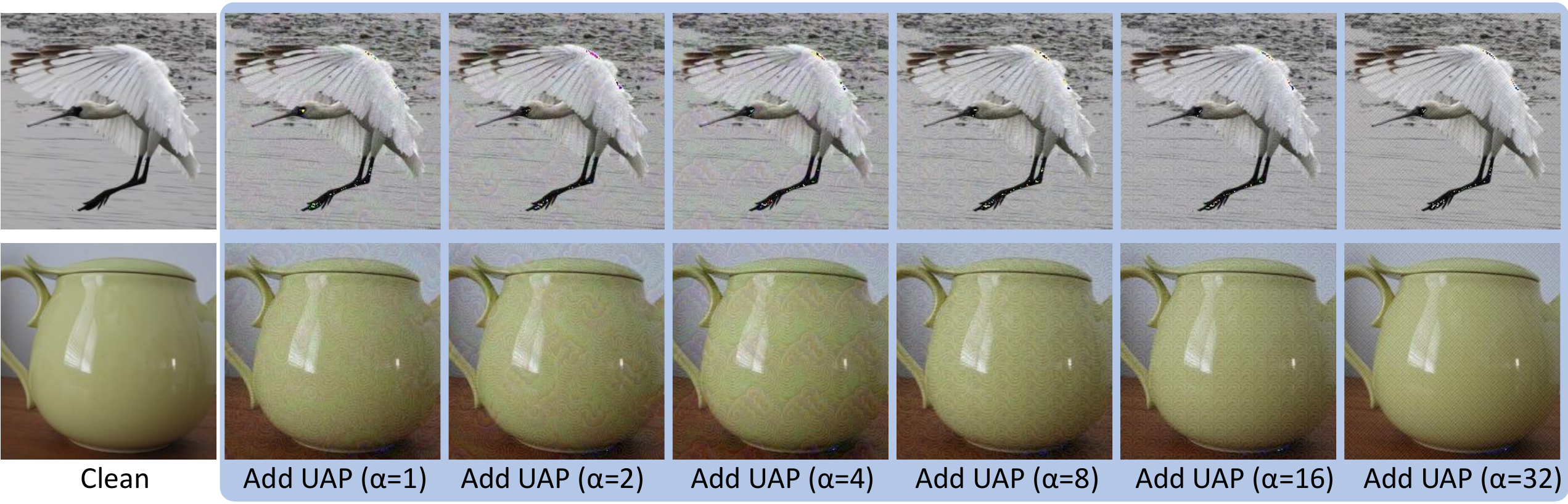}
\caption{Two clean images and their perturbed version by adding UAPs of different $\alpha$ on them.}
\label{fig:add_UAP_visualization}
\end{figure*}

We apply UAPs on clean images and visualize them in Figure~\ref{fig:add_UAP_visualization}. Here we randomly select two clean images from ImageNet and add UAPs to them. To be specific, we use the six UAPs (with different $\alpha$) generated with ResNet (\ie, the UAPs in the first row of Figure~\ref{fig:show_UAPs}). It is observed that these perturbations do not alter human classification ability, \ie, the crane and cups remain distinctly recognizable.

\section{Discussion}\label{sec:discussion}
\paragraph{Advantages of TSC-UAP} According to the experiment results, we can find that texture scale constraints have four advantages: \ding{182} Enhance fooling ratio: this is the main metric of UAP and TSC significantly improves the performance. \ding{183} Enhance attack transferability: TSC-UAP generates UAPs of higher cross-model transferability and cross-data transferability, which further enhance the black-box ability of UAP. \ding{184} Low cost: TSC is almost no extra overhead and may reduce the calculated amount since the update only applies to local texture, which is smaller than on the whole UAP. \ding{185} High generality: TSC can be utilized to improve both data-dependent and data-free UAP generation methods. Our work is a practical and preliminary endeavor intended to substantiate the significance of a new research direction in UAP generation through empirical study. The contribution is proposing the huge influence of texture on UAP generation and the proposed TSC is a simple and reasonable way to exploit the issue.

\paragraph{Tansferability between CNNs and ViT}
We conduct experiment of TSC-UAP on ViT \cite{dosovitskiy2021an} and evaluate the transferability between CNN and ViT in Table~\ref{tab:fr-CNN_ViT} and Table~\ref{tab:fr-ViT_CNN}. In Table~\ref{tab:fr-CNN_ViT}, we evaluate the transferability of UAPs generated by ResNet50, VGG19, DenseNet121 and MobileNet-v2 to the ViT-B/16 model. We can find that the attack performance of these UAPs is not good on ViT but TSC-UAP can still achieve higher fooling ratios than UAP without TSC constraint (\ie, $\alpha = 1$). In Table~\ref{tab:fr-ViT_CNN}, we evaluate the transferability of UAPs generated by ViT-B/16 to ResNet50, VGG19, DenseNet121 and MobileNet-v2 models. First, TSC-UAP is effective on ViT since when $\alpha = 2,4,8,16,32$, the fooling ratios are all higher than UAP without TSC constraint. Also, we can find that the attack performance of these UAPs is high on CNNs, even higher than that on ViT, which means attacking ViT to generate UAP is harder than attacking CNN.

\begin{table}[tb]
\center
\setlength{\tabcolsep}{2.5pt}
\caption{Transferability from UAPs generated by CNNs to ViT (\ie, ResNet50, VGG19, DenseNet121, MobileNet-v2, ViT-B/16) are in the second row and the texture scale parameter $\alpha$ is in the first column. Each cell shows the fooling ratio.}
\resizebox{\linewidth}{!}{
\begin{tabular}{c|cc|cc|cc|cc}
\hline 
\multirow{2}{*}{\diagbox{$\alpha$}{Model}} & Source & Target & Source & Target & Source & Target & Source & Target\tabularnewline
\cline{2-9}
 & ResNet50 & ViT-B/16 & VGG19 & ViT-B/16 & DenseNet121 & ViT-B/16 & MobileNet-v2 & ViT-B/16\tabularnewline
\hline 
1 & 80.30 & 17.34 & 82.44 & 16.97 & 66.05 & 18.96 & 94.08 & 14.20\tabularnewline
2 & 87.22 & 18.82 & 90.37 & 18.53 & 78.83 & 20.94 & 97.01 & 15.36\tabularnewline
4 & 89.92 & 20.99 & 92.88 & 20.74 & 82.26 & 22.22 & 97.25 & 16.49\tabularnewline
8 & 92.73 & 23.14 & 86.57 & 18.78 & 85.18 & 23.53 & 99.19 & 10.73\tabularnewline
16 & 78.47 & 16.97 & 83.54 & 18.42 & 79.26 & 20.49 & 97.53 & 10.18\tabularnewline
32 & 53.96 & 14.81 & 79.39/ & 17.31 & 51.24 & 23.87 & 74.65 & 22.30\tabularnewline
\hline 
\end{tabular}}
\label{tab:fr-CNN_ViT}
\end{table}

\begin{table}[tb]
\center
\caption{Transferability from UAPs generated by ViT to CNN (\ie, ResNet50, VGG19, DenseNet121, MobileNet-v2, ViT-B/16) are in the second row and the texture scale parameter $\alpha$ is in the first column. Each cell shows the fooling ratio.}
\resizebox{\linewidth}{!}{
\begin{tabular}{c|c|cccc}
\hline 
 \multirow{2}{*}{\diagbox{$\alpha$}{Model}} & Source & \multicolumn{4}{c}{Target}\tabularnewline
\cline{2-6} 
 & ViT-B/16 & ResNet50 & VGG19 & DenseNet121 & MobileNet-v2\tabularnewline
\hline 
1 & 21.43 & 21.69 & 31.05 & 22.03 & 31.65\tabularnewline
2 & 76.78 & 35.93 & 50.28 & 36.55 & 49.17\tabularnewline
4 & 42.63 & 43.82 & 55.60 & 43.80 & 55.78\tabularnewline
8 & 37.21 & 48.70 & 61.56 & 50.48 & 56.37\tabularnewline
16 & 31.32 & 34.99 & 57.19 & 35.83 & 55.75\tabularnewline
32 & 24.53 & 36.62 & 57.84 & 26.14 & 50.78\tabularnewline
\hline 
\end{tabular}}
\label{tab:fr-ViT_CNN}
\end{table}

\begin{figure}[tbp]
    \centering
    \includegraphics[width=0.8\linewidth]{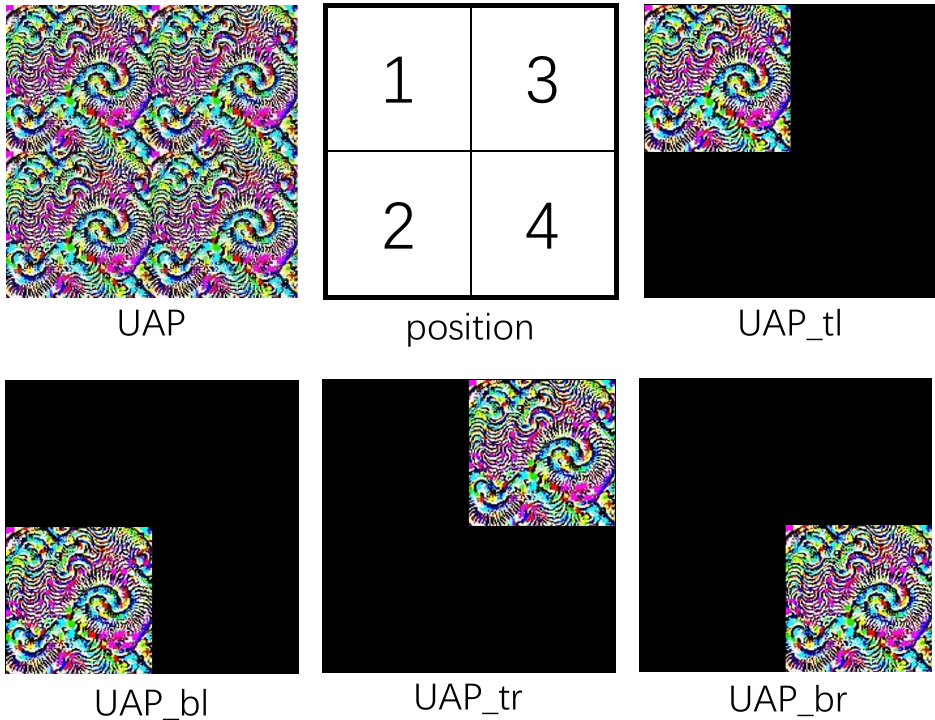}
    \caption{Influence of different UAP blocks (corners). Here we choose TSC-UAP with $\alpha$ = 2 and clip it into four UAP blocks.}
    \label{fig:UAP_patch_corner}
\end{figure}

\begin{figure}[tbp]
    \centering
    \includegraphics[width=0.8\linewidth]{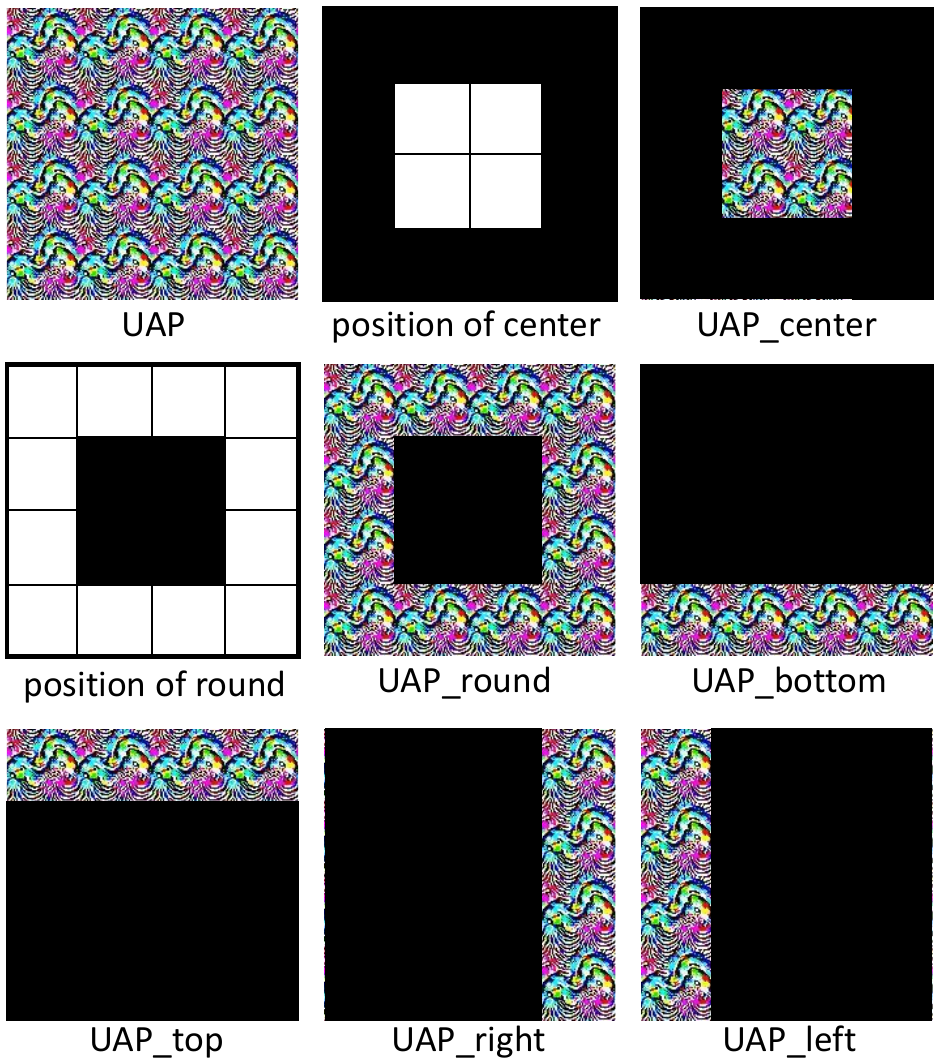}
    \caption{Influence of different UAP blocks (round and center). Here we choose TSC-UAP with $\alpha$ = 4 and clip it into different UAP blocks.}
    \label{fig:UAP_patch_round}
\end{figure}

\paragraph{Effect of UAP blocks}
The positions of the UAP patches are significant toward attack performance. Here we take TSC-UAP with different $\alpha$ as examples. In Figure~\ref{fig:UAP_patch_corner}, we use the TSC-UAP with $\alpha$ = 2 and generate four new UAPs (UAP\_tl, UAP\_bl, UAP\_tr, UAP\_br) which only reserve one of the four UAP patches. We use the same experimental setting as in Table 1 of our original paper except replacing UAP with new UAP. In Table~\ref{tab:fr-UAP-corner-round}, we can find that each patch achieves similar attack performance. We also conduct experiments on TSC-UAP with $\alpha$ = 4. In Figure~\ref{fig:UAP_patch_round}, we either reserve the center or the round of the UAP to generate some new UAPs. In Table~\ref{tab:fr-UAP-corner-round}, we can find that UAP with only center patches reserved achieves higher FR than around UAP patches which have the same perturbation quantity. We think this is because the objects in images usually exist in the center and UAP\_center is more likely to mask the object in the image, which makes the classifier focus on the UAP patch and misclassify the image.

\begin{table}[h]
\centering
\caption{Evaluation of UAP patches (different positions) on various CNNs.}
\resizebox{\linewidth}{!}{
\begin{tabular}{l|cccc}
\hline 
 FR (\%)& ResNet50 & VGG19 & DenseNet121 & MobileNet-v2\tabularnewline
\hline 
UAP ($\alpha$ = 2) & 87.22 & 90.37 & 78.83 & 97.01\tabularnewline
UAP\_tl ($\alpha$ = 2) & 23.28 & 36.88 & 21.67 & 31.42\tabularnewline
UAP\_bl ($\alpha$ = 2) & 19.88 & 31.10 & 19.52 & 29.89\tabularnewline
UAP\_tr ($\alpha$ = 2) & 21.30 & 37.39 & 21.60 & 30.28\tabularnewline
UAP\_br ($\alpha$ = 2) & 18.53 & 31.41 & 19.36 & 28.17\tabularnewline
\hline 
UAP ($\alpha$ = 4) & 89.92 & 92.88 & 82.26 & 97.25\tabularnewline
UAP\_center ($\alpha$ = 4)& 33.46 & 43.36 & 26.55 & 48.17\tabularnewline
UAP\_round ($\alpha$ = 4)& 75.20 & 81.18 & 65.53 & 79.09\tabularnewline
UAP\_top ($\alpha$ = 4)& 21.43 & 34.01 & 19.73 & 25.26\tabularnewline
UAP\_right ($\alpha$ = 4)& 23.07 & 30.75 & 19.70 & 28.18\tabularnewline
UAP\_bottom ($\alpha$ = 4)& 19.29 & 31.13 & 18.38 & 24.27\tabularnewline
UAP\_left ($\alpha$ = 4)& 23.55 & 33.53 & 20.08 & 30.65\tabularnewline
\hline 
\end{tabular}}
\label{tab:fr-UAP-corner-round}
\end{table}

\paragraph{TSC-UAP against defense method}
We conduct experiment against UAP defense methods \cite{akhtar2018defense}. It is the classical UAP defense method and according to our investigation, it is the only one that open-sources the code and provides the defense model. Please note that the defense model provided by \cite{akhtar2018defense} can choose to turn on or off the defense. Thus fooling ratios before/after defense are both achieved on the same defense model. We use the UAPs generated in Table~\ref{tab:fr-TSC-SGDUAP} as input since the UAPs with $\alpha = 1$ can represent the UAPs without TSC while the others are generated with our TSC-UAP. As shown in Table~\ref{tab:fr-TSC-SGDUAP_defense}, UAP source models (\ie, ResNet50, VGG19, DenseNet121, MobileNet-v2) are in the first row and the texture scale parameter $\alpha$ is in the first column. In each cell, the fooling ratios before/after defense are on the left/right and the decrement is in the parenthesis. We can observe two points. \ding{182} From the attack performance of UAPs on the model provided by \cite{akhtar2018defense} without defense, we can find that the TSC constraint is effective. \ding{183} We can find that for UAPs generated by the same source model with different $\alpha$, the decrement is similar. To summarize, the performance of TSC-UAP against the defense model is similar to the UAP without TSC constraint. However, we have to emphasize that TSC-UAP can achieve higher fooling ratios, thus with similar decrement, TSC-UAP can still achieve better attack performance than UAP without TSC constraint. Since TSC-UAP is not designed against defense methods, we think the result is reasonable and we will focus on designing the UAP generation method against the defense model in future work.

\begin{table}[htb]
\center
\setlength{\tabcolsep}{2.5pt}
\caption{For the performance of TSC-UAP against defense methods, UAP source models (\ie, ResNet50, VGG19, DenseNet121, MobileNet-v2) are in the first row and the texture scale parameter $\alpha$ is in the first column. In each cell, the fooling ratios before/after defense are on the left/right and the decrement is in the parenthesis.}
\resizebox{\linewidth}{!}{
\begin{tabular}{c|cccc}
\toprule 
\diagbox{$\alpha$}{Model} & ResNet50 & VGG19 & DenseNet121 & MobileNet-v2\tabularnewline
\hline 
1 & 40.05/30.02(10.03) & 33.60/27.32(6.28) & 43.08/35.27(7.81) & 30.38/22.65(7.73)\tabularnewline
2 & 45.02/36.51(8.51) & 41.05/31.64(9.41) & 51.01/41.67(9.34) & 34.44/25.46(8.98)\tabularnewline
4 & 45.62/37.08(8.54) & 46.92/35.32(11.60) & 52.57/43.92(8.65) & 39.03/27.88(11.15)\tabularnewline
8 & 63.80/41.36(22.44) & 43.34/34.76(8.58) & 62.57/51.18(11.39) & 23.97/19.29(4.68)\tabularnewline
16 & 48.54/36.37(12.17) & 35.49/29.47(6.02) & 59.76/46.66(13.10) & 21.74/18.88(2.86)\tabularnewline
32 & 33.63/22.47(11.16) & 35.90/26.04(9.86) & 57.41/50.92(6.49) & 44.67/32.45(12.22)\tabularnewline
\hline 
\end{tabular}}
\label{tab:fr-TSC-SGDUAP_defense}
\end{table}

\paragraph{Potential Application}
It would be very interesting if UAP could be applied to a potential application in other domains (\eg, adversarial patch, malware detection). Here we list some potential ideas for applying UAP in different domains. The HARP \cite{cai2023harp} is a newly proposed model-side defense method against adversarial patches by only inserting lightweight CNN modules into the pre-trained object detectors. Compared with the adversarial patches mentioned in the paper, we think UAP is different from the two points. \ding{182} The size of UAP (\ie, same size as the input image) is much bigger than the adversarial patch and would fully mask the object. \ding{183} The perturbation of UAP is not as big as that in the adversarial patch. It would be interesting to evaluate and enhance the robustness of object detectors against UAP. BagAmmo \cite{li2023black} is a newly proposed novel black-box attack method against Function Call Graph (FCG) based Android malware detection systems. They proposed a new malware manipulation called ``try-catch trap''. We think taking malware manipulation as a kind of perturbation, maybe designing a ``universal'' malware manipulation method that can be effective at any place of FCG is interesting and practical. \cite{maiorca2019towards} comprehensively survey malicious PDF detection in adversarial environments, inspiring us that maybe designing a ``universal'' content that can fool malicious PDF detectors is an interesting application.

\paragraph{Theoretical Analysis of TSC-UAP}
Taking UAP generation as a search problem, then the gradient for optimizing UAP is the guidance for the search. If directly optimizing all the pixels in UAP with gradient, the search space is too big. It is hard to force all the meaningless noises into category-specific textures and usually parts region of the UAP still exist meaningless noises (see images with $\alpha=1$ (without TSC constraint) in Figure~\ref{fig:show_human_understandable_UAPs}). By using split ratio $\alpha>1$ to reduce the region in UAP that needs optimization (see our method Figure~\ref{fig:framework-TSCUAP}), the difficulty of optimization is reduced (search space becomes small) since our method only needs to search category-specific texture in the local area. With generated category-specific texture in the local area, the problem becomes how to make it exist in the entire UAP. To address the problem, patch tiling is a good choice for this goal. However, please note that patch tiling is not the only choice since other operations (\eg, flip) can also achieve the goal of making category-specific texture exist in the entire UAP and thus enhance UAP.

\paragraph{Limitations of TSC-UAP}
Although we point out the huge influence of texture on the attack performance of UAP, the way of exploiting such an issue may not be most appropriate. It would be better to give a theoretical analysis that can further promote the development of research on how to exploit the texture of UAP. However, we firmly believe that this practical study is essential and serves as a valuable starting point for theoretical analysis. Given the complexity of the matter, which cannot be fully elucidated by solely focusing on texture scale since other factors (\eg, texture category, the architecture of the target model, and training dataset) are also important, we consider theoretical analysis needs more observations and more appropriate for the future work.


\section{Conclusion}\label{sec:conclusion}
In this paper, we research the textures in universal adversarial perturbation and propose texture scale constraints to improve the UAP methods. The proposed texture scale constraints not only achieve a higher fooling ratio but also higher attack transferability with minor computational costs. It is also general enough to be applied to both data-dependent and data-free UAP methods. In future work, we aim to study the relationship between the texture scale and the receptive field of CNNs, which may be instructive for model explainability.

\section*{Appendix}
\subsection{Visualization of Untargeted UAPs across Different Models}
\begin{figure}[tbp]
    \centering
    \includegraphics[width=\linewidth]{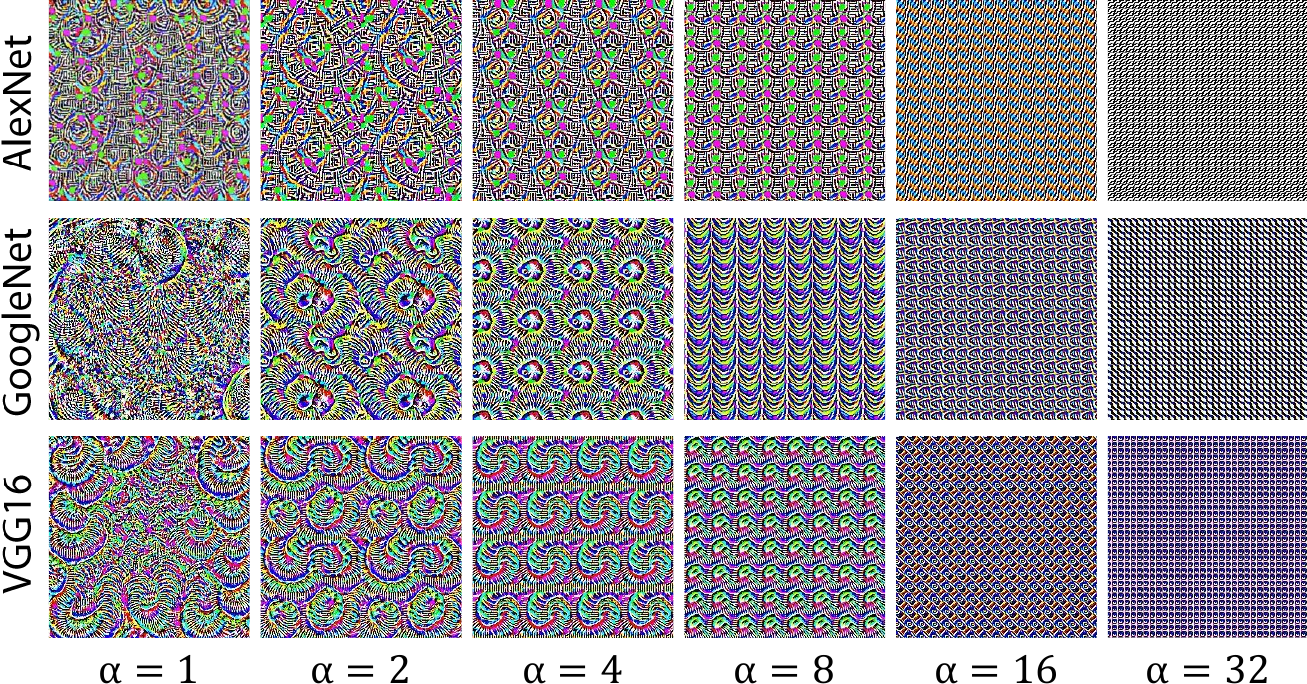}
    \caption{Here shows the UAPs generated by attacking Alexnet, GoogleNet and VGG16 on ImageNet training samples.}
    \label{fig:show_UAPs_supp_untarget}
\end{figure}

We show the untargeted UAPs on more CNNs (\ie, AlexNet, GoogleNet, VGG16) in Figure~\ref{fig:show_UAPs_supp_untarget}. We can find that, the texture on different models all seems special when $\alpha$ is not big and shows some semantic in the texture.

\subsection{Visualization of Targeted UAPs across Different Models}
\begin{figure}[tbp]
    \centering
    \includegraphics[width=\linewidth]{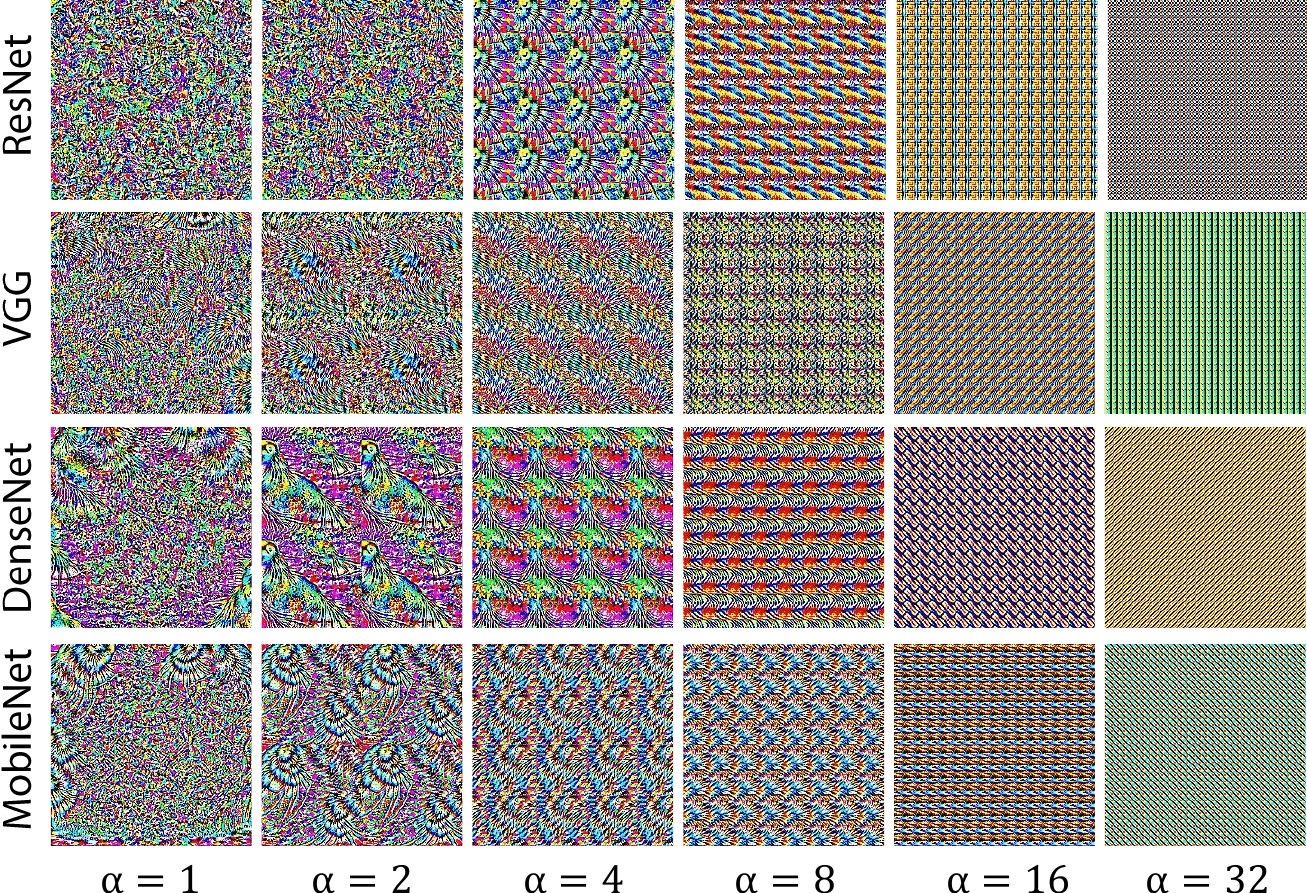}
    \caption{Here shows the UAPs generated by attacking ResNet50, VGG19, DenseNet121 and MobileNet-v2 on ImageNet training samples with target label sea lion.}
    \label{fig:show_UAPs_supp_target_sealion}
\end{figure}

\begin{figure}[tbp]
    \centering
    \includegraphics[width=\linewidth]{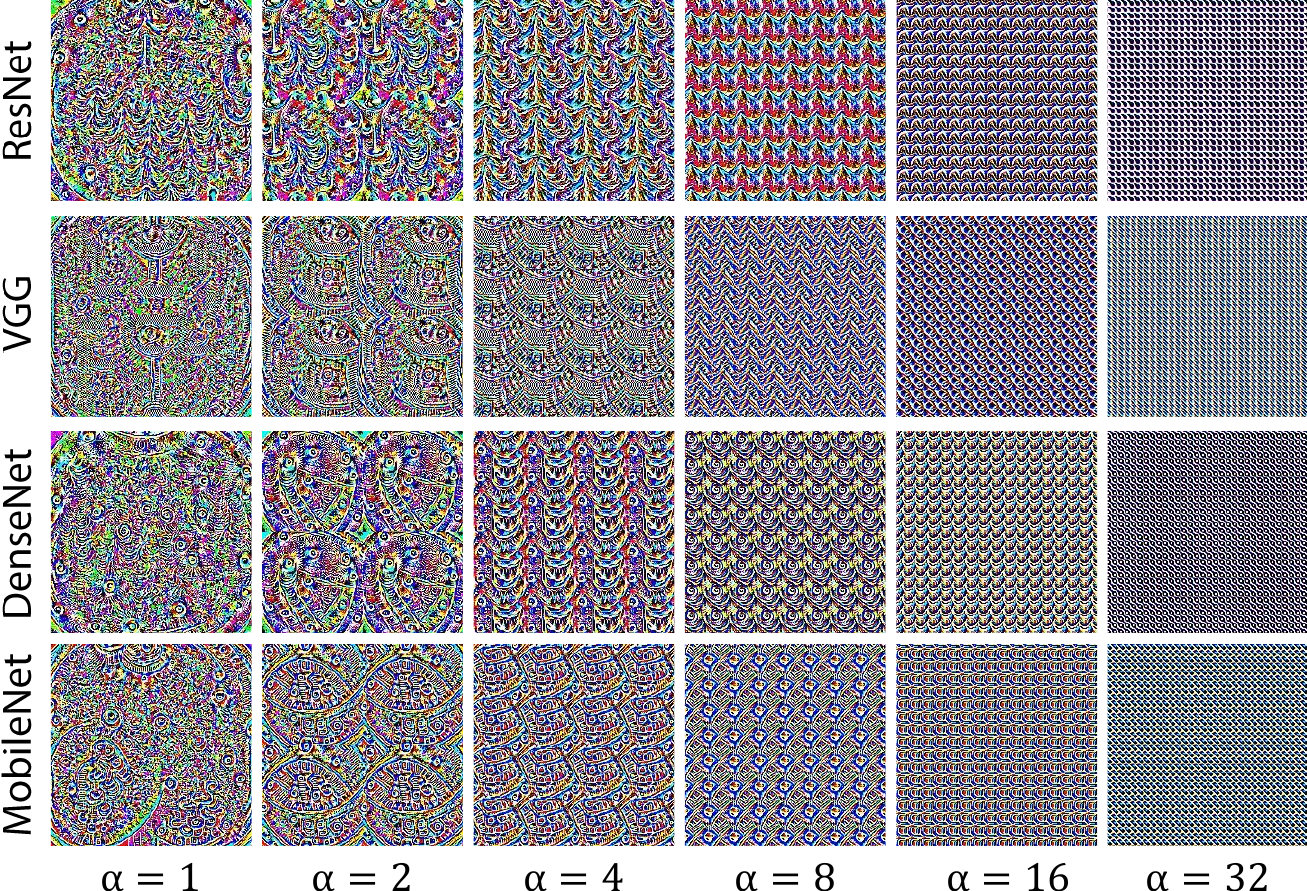}
    \caption{Here shows the UAPs generated by attacking ResNet50, VGG19, DenseNet121 and MobileNet-v2 on ImageNet training samples with target label shield.}
    \label{fig:show_UAPs_supp_target_shield}
\end{figure}

We show the targeted UAPs in Figure~\ref{fig:show_UAPs_supp_target_sealion} and Figure~\ref{fig:show_UAPs_supp_target_shield} by attacking ResNet50, VGG19, DenseNet121, and MobileNet-v2 on ImageNet training samples with randomly chosen target labels (sea lion and shield) respectively. We can find that there exist texture of sea lion and shield in Figure~\ref{fig:show_UAPs_supp_target_sealion} and \ref{fig:show_UAPs_supp_target_shield} respectively, especially when $\alpha$ is 2.


\bibliographystyle{IEEEtran}
\bibliography{ref}

\end{document}